\documentclass[letterpaper]{article} 
\usepackage{aaai2026}  

\usepackage{times}  
\usepackage{helvet}  
\usepackage{courier}  
\usepackage[hyphens]{url}  
\usepackage{graphicx} 
\usepackage{natbib}  
\usepackage{caption} 
\usepackage{algorithm}
\usepackage{algorithmic}
\usepackage{multirow}
\usepackage{booktabs}
\usepackage{makecell}
\usepackage{amsmath}
\usepackage{adjustbox}
\usepackage{newfloat}
\usepackage{listings}
\DeclareCaptionStyle{ruled}{labelfont=normalfont,labelsep=colon,strut=off} 
\floatstyle{ruled}
\newfloat{listing}{tb}{lst}{}
\floatname{listing}{Listing}

\usepackage{pifont}
\newcommand{\cmark}{\ding{51}} 
\newcommand{\xmark}{\ding{55}} 

\usepackage{ifthen}
\usepackage[table,dvipsnames]{xcolor}
\usepackage{enumitem}

\title{CogBench: A Large Language Model Benchmark for Multilingual Speech-Based Cognitive Impairment Assessment}

\author {
    Rui Feng\textsuperscript{\rm 1*},
    Zhiyao Luo\textsuperscript{\rm 2*\dag},
    Wei Wang\textsuperscript{\rm 1},
    Yuting Song\textsuperscript{\rm 3},
    Yong Liu\textsuperscript{\rm 3},
    Tingting Zhu\textsuperscript{\rm 2},
    Jianqing Li\textsuperscript{\rm 1\dag},
    Xingyao Wang\textsuperscript{\rm 3}
}
\affiliations {
    \textsuperscript{\rm 1}Engineering Research Center of Intelligent Theranostics Technology and Instruments, Ministry of Education, School of Biomedical Engineering and Informatics, Nanjing Medical University, Nanjing 211166, China\\
    \textsuperscript{\rm 2}Institute of Biomedical Engineering, University of Oxford, Oxford OX1 2JD, United Kingdom\\
    \textsuperscript{\rm 3}Institute of High Performance Computing (IHPC), Agency for Science, Technology and Research (A*STAR), Singapore 138632, Singapore\\
    bmefr@stu.njmu.edu.cn, \{zhiyao.luo, tingting.zhu\}@eng.ox.ac.uk, \{bmeww, jqli\}@njmu.edu.cn,\\
    \{songy, liuyong, wang\_xingyao\}@a-star.edu.sg
}

\usepackage{bibentry}

\nocopyright

\begin{document}

\maketitle

\let\thefootnote\relax
\footnotetext{* Equal contribution.}
\footnotetext{\dag\ Corresponding authors.}
\begin{abstract}

Automatic assessment of cognitive impairment from spontaneous speech offers a promising, non-invasive avenue for early cognitive screening. However, current approaches often lack generalizability when deployed across different languages and clinical settings, limiting their practical utility. In this study, we propose \textbf{CogBench}, the first benchmark designed to evaluate the cross-lingual and cross-site generalizability of large language models (LLMs) for speech-based cognitive impairment assessment. Using a unified multimodal pipeline, we evaluate model performance on three speech datasets spanning English and Mandarin: ADReSSo, NCMMSC2021-AD, and a newly collected test set, CIR-E. Our results show that conventional deep learning models degrade substantially when transferred across domains. In contrast, LLMs equipped with chain-of-thought prompting demonstrate better adaptability, though their performance remains sensitive to prompt design. Furthermore, we explore lightweight fine-tuning of LLMs via Low-Rank Adaptation (LoRA), which significantly improves generalization in target domains. These findings offer a critical step toward building clinically useful and linguistically robust speech-based cognitive assessment tools.

\end{abstract}

\section{Introduction}

The global population’s rapid aging drives urgent demand for scalable, affordable early cognitive impairment detection~\cite{collaborators2019global}. As such, language analysis offers promise as a non-invasive screening modality~\cite{fristed2022leveraging,garcia2024unveiling}. However, current language assessments typically require structured administration by trained clinicians, limiting their feasibility for routine screening, particularly in resource-constrained settings and among older adults with low engagement or limited access to care~\cite{dokholyan2022challenges}.

Recent advancements in machine learning have demonstrated promising accuracy in detecting cognitive impairments through language analysis. The sensitivity of picture description tasks to such impairments makes them an ideal assessment tool for any investigation aiming to identify pragmatic markers of neurodegenerative diseases like dementia. AI-based approaches aim to predict cognitive status automatically using the raw audio signal, therefore bypassing the need for manual scoring. These AI methods have been enabled by widely used datasets such as Pitt~\cite{becker1994natural}, ADReSS~\cite{LuzHaiderEtAl20ADReSS}, and ADReSSo~\cite{LuzEtAl21ADReSSo}. Despite encouraging results, current models often fail to generalise across clinical settings and diverse populations, limiting their broader applicability~\cite{liu2021automatic,runde2024optimization}.

Parallel advances in decoder-only LLMs have revealed strong multimodal reasoning in complex medical tasks such as diagnosis~\cite{liu2025generalist} and report generation~\cite{alkhaldi2024minigpt}. Compared to small deep learning models, LLMs offer stronger generalization and interpretability, making them more suitable for clinical deployment in many clinical scenarios. Although some studies have begun to explore the use of LLMs in the cognitive domain, such as the work of Mo et al.~\cite{mo2025dect} using unstructured audio transcripts to extract language markers, the effectiveness of LLMs as screening tools for cognitive impairment remains largely unexplored.

In response to these challenges, this study aims to address three main objectives. First, we seek to build the first unified arena in automatic cognitive function assessment where different models can be evaluated using standardised protocols. Second, we investigate generalisation across different languages and datasets, an essential yet unexplored aspect of AI-based cognitive assessment. Third, we explore the potential of multimodal large language models (MLLMs) in this domain, aiming to determine whether they can outperform small-scale models (SSMs).

Our contributions are outlined as follows:
\begin{enumerate}

\item We provide CIR-E, a new Mandarin dataset from naturalistic community settings, to support linguistically and demographically diverse cognitive assessment research.

\item We present the first cross-lingual and cross-site benchmark for speech-based cognitive assessment, combining two public Chinese and English datasets with an extra test set, CIR-E. The benchmark supports comprehensive evaluation of representative SSMs and MLLMs.

\item We investigate the application of MLLMs through systematic prompt engineering, evaluating zero-shot, expert-knowledge (EXP), and chain-of-thought (CoT) prompting strategies with majority voting, benchmarked against time-domain and frequency-domain SSMs.

\item We compare domain adaptation performance between SSMs and MLLMs in a specialized domain, demonstrating that fine-tuning MLLMs exhibit superior generalization capabilities.

\end{enumerate}

To our knowledge, this study is the first to demonstrate that MLLMs can effectively serve as universal cognitive impairment screeners. We openly release our datasets, code, and evaluation scripts to encourage future research and promote equitable advances in global cognitive health assessment.

\section{Related Work}
\subsection{AI for Speech-Based Cognitive Assessment}

Numerous studies have explored speech-based AI methods for detecting cognitive impairment, demonstrating promising results across various datasets, including deep learning methods such as CNNs~\cite{ding2024speech}, LSTMs~\cite{igarashi2022cognitive}, and more recent multimodal~\cite{rohanian2021alzheimer}, ensemble~\cite{alkenani2021predicting}, and transfer learning frameworks~\cite{yang2022augmented}. However, the field lacks consistency in experimental settings, with studies often using different datasets, metrics, and protocols, making it difficult to compare methods fairly or draw generalizable conclusions. Most work focuses on optimizing performance within a single dataset, leaving open questions about how well these models generalize across tasks, speakers, and recording conditions.

\subsection{Large Language Models in Medical Applications}

LLMs have demonstrated strong potential across a range of medical domains. In pathology, CHIEF~\cite{wang2024pathology} achieves high accuracy in cancer diagnosis; in dermatology, SkinGPT-4~\cite{zhou2024pre} enables interactive diagnoses from skin images; in drug discovery, TxGNN~\cite{huang2024foundation} facilitates knowledge-based drug repurposing; and in genomics, DNABERT-2~\cite{kabir2024dna} improves the prediction of transcription factor binding sites. These examples highlight the impact of LLMs in clinical applications, especially when paired with domain supervision or multimodal input. However, their use in cognitive assessment remains underexplored. This work addresses that gap by evaluating and enhancing LLMs for speech-based cognitive screening.

\section{Methodology}

\subsection{Datasets}

For a comprehensive cross-site and cross-language evaluation, our study utilizes two public datasets and one external clinical dataset, covering both English and Mandarin across binary and ternary classification tasks. As shown in Table~\ref{tab:datasets}, The datasets include: 
1)~\textbf{ADReSSo}~\cite{LuzEtAl21ADReSSo}, a English corpus for Alzheimer’s Disease (AD) classification introduced at INTERSPEECH 2021. Participants were asked to describe the \textit{``Cookie Theft''} picture from the Boston Diagnostic Aphasia Examination, eliciting spontaneous speech. 
2)~\textbf{NCMMSC2021}~\cite{ChenXC2023}, a Mandarin corpus from the NCMMSC2021 Challenge for classification among AD, mild cognitive impairment (MCI), and healthy controls (HC), which includes multiple cognitive assessment tasks, such as picture description, fluency tests, and self-introductions. 
3)~\textbf{CIR-E}, an external Mandarin dataset collected from community participants in a real-world clinical setting in Jiangsu Province, China, with speech recordings evaluated by neurologists using standardized cognitive assessments. Group labels were assigned based on clinical evaluation, and the dataset was balanced for age and gender; it follows the same ternary classification scheme as NCMMSC2021.

We implemented a unified preprocessing pipeline to ensure consistent data quality across datasets. First, participant speech is isolated from dialogues using \textit{speaker-diarization-3.1}. The cleaned audio segments are then transcribed using \textit{faster-whisper}. Each sample is ultimately represented as a multimodal pair $(a_i, t_i)$, consisting of the participant's speech and corresponding transcript.

\begin{table}[t]
\centering
\small
\begin{tabular}{llcc}
\toprule
\textbf{Dataset} & \textbf{Task}& \textbf{Train / Test} & \textbf{Language} \\
\midrule
ADReSSo & Pic. Desc.  & 166 / 71  & English \\
\midrule
\multirow{3}{*}{NCMMSC2021} & Pic. Desc. & \multirow{3}{*}{280 / 119}  & \multirow{3}{*}{Chinese}  \\
& Fluency & & \\
& Self-intro. & & \\
\midrule
CIR-E (\textbf{Ours}) & Pic. desc.
 & -- / 153    & Chinese \\
\bottomrule
\end{tabular}
\caption{Overview of datasets used for cognitive assessment in this study. 'Pic. Desc.' denotes picture description. 'Fluency' denotes fluency tests. 'Self-intro.' denotes Self-introduction. 'Train / Test' denotes the number of patients in the training and test sets.}
\label{tab:datasets}
\end{table}

\begin{figure*}[t]
    \centering
    \includegraphics[width=0.95\linewidth]{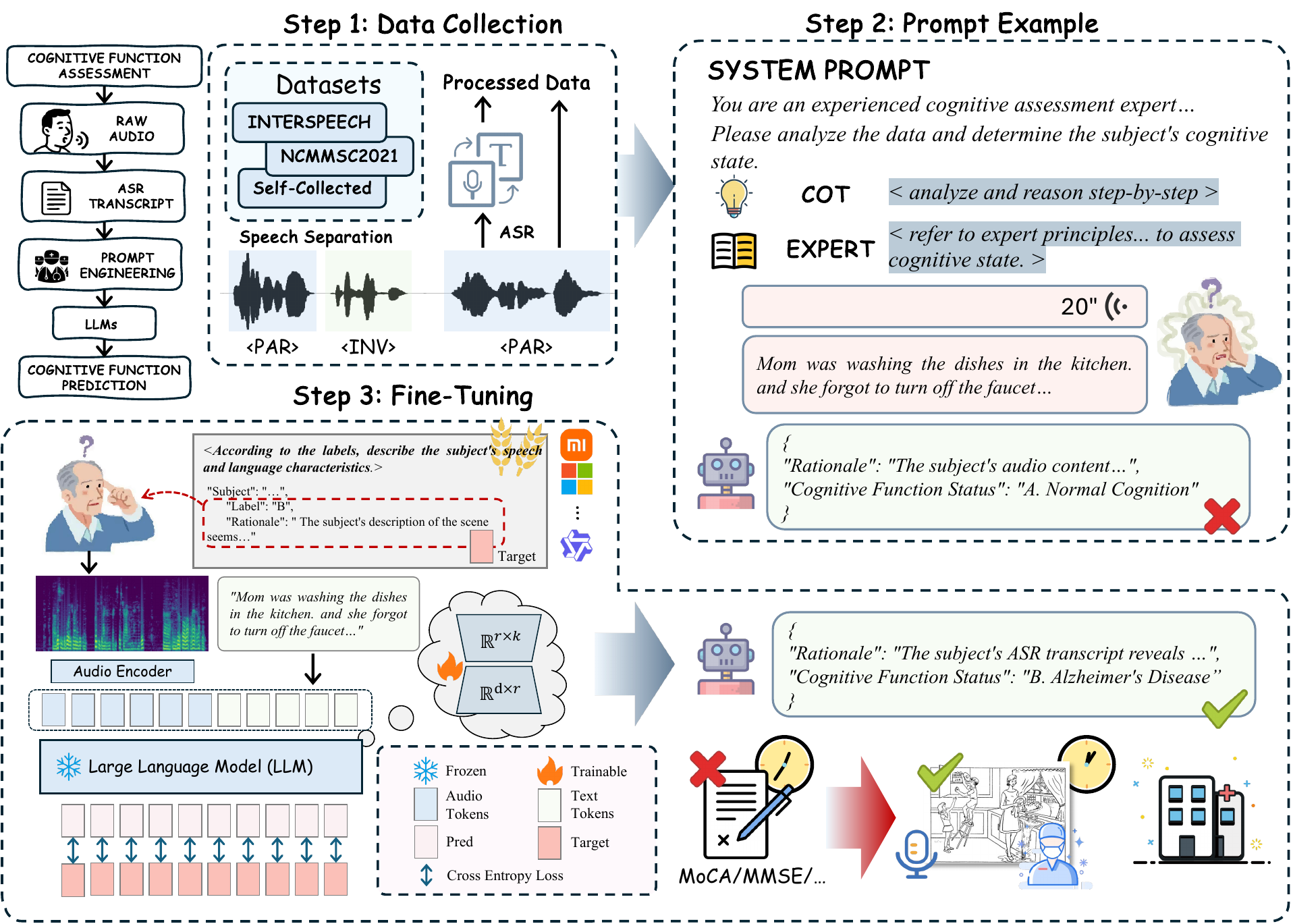}
    \caption{The overall workflow of our approach includes three key steps: \textbf{(1)} data preprocessing—performing speaker diarization and ASR to produce clean audio and transcripts from multiple datasets; \textbf{(2)} applying designed prompts to LLMs for cognitive status inference from multimodal inputs; and \textbf{(3)} fine-tuning LLMs via Low-Rank Adaptation (LoRA) on formatted data, followed by evaluation of the fine-tuned models to obtain final predictions.}

    \label{fig:workflow}
\end{figure*}

\subsection{Problem Definition}

We formalize the cognitive impairment assessment task as a supervised classification problem. Let $\mathcal{D} = \{(x_i, y_i)\}_{i=1}^N$ be a dataset of $N$ participants. Each instance $x_i$ is a multimodal sample, $x_i = (a_i, t_i)$, composed of a raw audio signal $a_i \in \mathcal{A}$ and its corresponding Automatic Speech Recognition (ASR) transcript $t_i \in \mathcal{T}$. The label $y_i \in \mathcal{Y}$ represents the participant's clinically-validated cognitive status, typically determined by standardized cognitive scale. For a binary classification task, the label set is $\mathcal{Y} = \{\textit{AD}, \textit{Non-AD}\}$, while for a tertiary task, it is $\mathcal{Y} = \{\textit{AD}, \textit{MCI}, \textit{HC}\}$. Our primary objective is to learn a mapping function $f: \mathcal{A} \times \mathcal{T} \to \mathcal{Y}$ that accurately predicts the cognitive label $\hat{y} = f(x)$ for an unseen sample $x$.

To adapt this classification task to LLMs, we recast it as a structured text generation problem. Given a prompt template $P$ embedding the sample $(a_i, t_i)$, the model $\mathcal{M}$ generates a response $S_{\text{gen}} = \mathcal{M}(P(x_i))$. The output is constrained to a JSON format:
{\small
\begin{verbatim}
{
  "Rationale": "<Reasoning process>",
  "Cognitive Function Status": "<Option>"
}
\end{verbatim}
}
The final predicted class, $\hat{y}$, is then programmatically extracted from the \texttt{<Option>} field of the generated JSON.

Figure~\ref{fig:workflow} illustrates the overall workflow of our approach.

\section{Experiments and Results}

This section presents our experimental results on cross-lingual and cross-site cognitive impairment classification. We systematically evaluate both SSMs and LLMs to understand their capabilities under different settings. Our results are organized around three core research questions: \\
\textbf{RQ1: \textit{How well do small models generalize across datasets and languages?}}\\
As discussed in the \textbf{OOD Evaluation} and \textbf{Linear Probing} sections, we investigate the generalization ability of SSMs  across languages and datasets, highlighting the challenges posed by domain shift for SSMs. \\
\textbf{RQ2: \textit{Can off-the-shelf LLMs effectively conduct speech-based cognitive impairment assessment with prompt engineering?}} \\
We evaluate the zero-shot performance of LLMs on multilingual speech tasks under different prompt strategies, focusing on their potential as universal assessors. \\
\textbf{RQ3: \textit{With fine-tuning, can LLMs beat SSMs?}}\\
In the \textbf{Comparing the Performance of SSMs and LLMs} section, we apply LoRA-based fine-tuning with domain-specific training samples and compare their adapted performance against both zero-shot LLMs and supervised SSMs. Meanwhile, the \textbf{Test-Time Scaling} section evaluates how inference-time scaling impacts model robustness, and the \textbf{Case Study} section analyzes representative failure cases.

\subsection{Experimental Setup}
\label{setup}

For SSMs training, we used the Adam optimizer with a cosine learning rate decay schedule. Hyperparameters such as learning rate and batch size were tuned through grid search to achieve optimal performance. Given the variability in recording lengths, all audio data were segmented using a 6s sliding window with a 2s stride. During inference, predictions across segments were aggregated via majority voting.

For LLMs experiments, we set the \textit{temperature} to 0.7 and \textit{top\_p} to 1.0, generating \textit{n\_sample} = 5 outputs per prompt to support sampling-based reasoning. To ensure fairness and robustness, all baseline results were averaged over five independent runs. Additional implementation details and parameter configurations are provided in the Appendix.

\subsection{Evaluation of Generalization for SSMs}
\label{acoustic_model_performance}

To establish a supervised learning baseline, we trained four SSMs: two time-domain models operating directly on raw audio waveforms—\textit{1D-ResNet} and \textit{LSTM}—and two frequency-domain models using mel-spectrogram features—\textit{ResNet18} and \textit{Transformer}.

\begin{table}[t]
\centering
\small
\begin{tabular}{>{\centering\arraybackslash}p{0.7cm}|c|l|>{\centering\arraybackslash}p{1cm}>{\centering\arraybackslash}p{1cm}>{\centering\arraybackslash}p{1cm}}

\toprule
  \multirow{2}{*}{\textbf{Train}}   & \multirow{2}{*}{\textbf{cls}}   & \multirow{2}{*}{\textbf{Model}}  & \multicolumn{3}{c}{\textbf{Test}} \\
       &     &    & ADReSSo & NCMMSC & CIR-E  \\
\midrule
\multirow{4}{*}{\rotatebox[origin=c]{90}{ADReSSo}} 
    & \multirow{4}{*}{2}    & 1D-ResNet   & \cellcolor{gray!20} \textbf{63.37} & \textbf{59.84} & \textbf{53.00}  \\
   &  & LSTM        & \cellcolor{gray!20} 54.82 & 43.30 & \underline{29.18} \\
   &  & ResNet18    & \cellcolor{gray!20} \underline{63.20}  & 26.14 & 23.12   \\
   &  & Transformer & \cellcolor{gray!20} 60.49 & \underline{51.61} & 23.12 \\
   
\midrule
\multirow{8}{*}{\rotatebox[origin=c]{90}{NCMMSC2021}}  
      & \multirow{4}{*}{2}  & 1D-ResNet   & \underline{36.74} &  \cellcolor{gray!20} 81.97  & 45.01  \\
    & & LSTM        & \textbf{39.69} &  \cellcolor{gray!20} \textbf{87.16} & \underline{48.73} \\
    & & ResNet18    & 33.64 &  \cellcolor{gray!20} \underline{86.06} & \textbf{52.96} \\
    & & Transformer & 32.38 &  \cellcolor{gray!20} 71.16 & 47.76 \\
\cmidrule(lr){2-6}
   & \multirow{4}{*}{3}   & 1D-ResNet   & / & \cellcolor{gray!20} 74.52 & \underline{21.91} \\
   &  & LSTM        & / &  \cellcolor{gray!20} \textbf{85.64} & \textbf{24.27}\\
   &  & ResNet18    & / &  \cellcolor{gray!20} 80.36 & 17.78  \\
   &  & Transformer & / &  \cellcolor{gray!20} \underline{81.55} & 21.73  \\

\bottomrule
\end{tabular}
\caption{Model performance on ID and OOD data under label unification (Macro-F1 \%).}
\label{tab:Label_Unification}
\end{table}

\subsubsection{OOD Evaluation for SSMs}
\label{experiment1}

Following a binary label scheme, we merged the HC and MCI classes into a single Non-AD category to unify labels with the ADReSSo dataset. Table~\ref{tab:Label_Unification} summarizes SSMs’ performance under unified label settings across binary and ternary classification tasks in both in-domain (ID) and out-of-domain (OOD) scenarios. A clear performance drop in OOD settings reveals limited model generalization under domain shifts.
Models trained on NCMMSC2021 generally outperform those trained on ADReSSo, particularly in OOD evaluations. This may be due to differences in dataset construction: ADReSSo is explicitly balanced by age and gender, potentially reducing class-separable signals, whereas NCMMSC2021's distribution may inherently offer stronger discriminative cues.
Label granularity also influences performance. Reducing the original ternary classification to binary improves results in some cases but introduces fluctuations in F1 scores. As a transitional cognitive state, MCI exhibits high variability across datasets, making its boundaries harder to define.
Cross-lingual transfer results further underscore the role of linguistic and demographic alignment. Models trained on the Mandarin-language NCMMSC2021, which is closer to CIR-E in both language and population, transfer better than those trained on the English-language ADReSSo dataset.

Overall, these findings highlight the challenges of site and language mismatch. Simple label unification alone is insufficient to address cross-dataset discrepancies.

\begin{table}[t]
\centering
\small
\begin{tabular}{p{2.2cm}>{\centering\arraybackslash}p{2cm}cc}
\toprule
\multirow{2}{*}{\textbf{Model}} & \multicolumn{2}{c}{\textbf{Domain Adaptation}} & \multirow{2}{*}{\textbf{ID}} \\
                                & \makecell[c]{w / o $\rightarrow$  w / LP}  & $\Delta$ (\%) & \\
\midrule
\multicolumn{4}{l}{\textbf{ADReSSo $\rightarrow$ NCMMSC2021}} \\
\midrule
1D-ResNet          & 59.84 $\rightarrow$ 42.15 &  $-17.69$  & 74.52   \\
LSTM               & 43.30 $\rightarrow$ 32.77 &  $-10.53$  &  85.64 \\
ResNet18           & 26.14 $\rightarrow$ 36.43 &  $\mathbf{+10.29}$  &  80.36 \\
Transformer        & 51.61 $\rightarrow$ 30.44 &  $-21.17$  & 81.55 \\

\midrule
\multicolumn{4}{l}{\textbf{NCMMSC2021 $\rightarrow$ ADReSSo}} \\
\midrule
1D-ResNet          & 36.74 $\rightarrow$ 58.63 &  $\mathbf{+21.89}$ &  63.37  \\
LSTM               & 39.69 $\rightarrow$ 38.92 &  $-0.77$  &  54.82  \\
ResNet18           & 33.64 $\rightarrow$ 33.02 &  $-0.62$ &  63.20  \\
Transformer        & 32.38 $\rightarrow$ 33.02 &  $\mathbf{+0.64}$ &  60.49  \\
 
\bottomrule
\end{tabular}
\caption{Performance of models on domain adaptation tasks with linear probing (Macro-F1 \%).}
\label{tab:linearprobing}
\end{table}

\subsubsection{Linear Probing}
\label{experiment2}

Given the failure of direct transfer, we then explored a simple and efficient domain adaptation strategy, linear probing. Specifically, we treat the pre-trained model on the source domain as a fixed feature extractor, freeze most of its network layers, and then fine-tune its final classification layer only on a small amount of training data from the target domain. This experiment aims to verify whether the deep features extracted by the model have certain transfer value.

Table~\ref{tab:linearprobing} shows that linear probing does not consistently improve performance across domains. In several cases, particularly in cross-lingual settings such as ADReSSo to NCMMSC2021, linear probing yields noticeably lower macro-F1 scores compared to direct transfer under label unification. This suggests that audio representations learned in the source domain may not transfer effectively to the target domain, especially when there is a language mismatch.
The inconsistency is likely due to intrinsic differences in pronunciation, prosody, and cognitive expression patterns between languages. These variations affect the acoustic and semantic distribution of speech features, leading feature extractors trained on a single domain to specialize in domain-specific characteristics, which in turn restricts their generalizability to other domains.

\begin{table*}[t]
\centering
\small
\begin{tabular}{l|cc|cc|cc|cc}
\toprule
\multirow{2}{*}{\textbf{Model}} & \multicolumn{2}{c|}{\textbf{Prompt}} & \multicolumn{2}{c|}{\textbf{ADReSSo}} & \multicolumn{2}{c|}{\textbf{NCMMSC2021}} & \multicolumn{2}{c}{\textbf{CIR-E}} \\
 & COT & EXP & Maj@5 & Avg@1 & Maj@5 & Avg@1 & Maj@5 & Avg@1 \\
\midrule
\multirow{4}{*}{\rotatebox[origin=c]{45}{R1}} 
 & \xmark & \xmark & 47.61${\pm6.15}$ & 42.57${\pm4.12}$ & 16.86${\pm0.70}$ & 16.00${\pm1.20}$ & 21.79${\pm0.12}$ & 21.60${\pm0.33}$ \\
 & \cmark & \xmark & 54.59${\pm3.18}$ & 53.90${\pm8.18}$ & 19.04${\pm1.79}$ & 21.47${\pm1.26}$ & 27.44${\pm2.42}$ & 29.86${\pm2.87}$ \\
 & \xmark & \cmark & 33.02${\pm0.00}$ & 33.62${\pm1.21}$ & 17.05${\pm0.81}$ & 17.31${\pm0.97}$ & 22.44${\pm1.42}$ & 22.41${\pm1.32}$ \\
 & \cmark & \cmark & 33.02${\pm0.00}$ & 32.89${\pm0.26}$ & 26.35${\pm3.22}$ & 28.19${\pm6.49}$ & 28.03${\pm2.81}$ & 28.37${\pm2.39}$ \\
\cmidrule(lr){1-9}
\multirow{4}{*}{\rotatebox[origin=c]{45}{Ultravox}} 
 & \xmark & \xmark & 40.62${\pm0.91}$ & 43.98${\pm7.95}$ & 22.04${\pm3.17}$ & 25.36${\pm7.64}$ & 27.67${\pm1.39}$ & 31.10${\pm2.50}$ \\
 & \cmark & \xmark & 41.56${\pm4.68}$ & 46.77${\pm4.77}$ & 19.21${\pm2.13}$ & 24.18${\pm1.83}$ & 22.08${\pm0.88}$ & 29.76${\pm6.93}$ \\
 & \xmark & \cmark & 34.80${\pm2.37}$ & 39.61${\pm4.81}$ & 25.87${\pm5.79}$ & 22.46${\pm7.77}$ & 25.81${\pm0.61}$ & 27.00${\pm3.38}$ \\
 & \cmark & \cmark & 41.65${\pm4.81}$ & 46.98${\pm2.98}$ & 29.16${\pm2.41}$ & 29.88${\pm5.86}$ & \textbf{\boldmath 31.84${\pm2.14}$} & \textbf{\boldmath 32.46${\pm2.62}$} \\
\cmidrule(lr){1-9}
\multirow{4}{*}{\rotatebox[origin=c]{45}{SeaLLMs}} 
 & \xmark & \xmark & 40.32${\pm9.14}$ & 40.08${\pm8.22}$ & 24.36${\pm5.66}$ & 21.08${\pm5.75}$ & 25.12${\pm6.21}$ & 23.58${\pm5.55}$ \\
 & \cmark & \xmark & 52.56${\pm2.85}$ & 48.14${\pm7.58}$ & \textbf{\boldmath 30.02${\pm2.81}$} & 27.89${\pm5.61}$ & 28.89${\pm2.01}$ & 30.08${\pm1.05}$ \\
 & \xmark & \cmark & 33.62${\pm1.21}$ & 35.34${\pm3.35}$ & 16.51${\pm0.91}$ & 17.19${\pm1.48}$ & 22.04${\pm0.63}$ & 22.43${\pm1.40}$ \\
 & \cmark & \cmark & 48.49${\pm10.99}$ & 42.38${\pm4.97}$ & 27.15${\pm3.85}$ & \textbf{\boldmath 31.84${\pm3.39}$} & 30.22${\pm3.28}$ & \underline{32.00${\pm6.29}$} \\
\cmidrule(lr){1-9}
\multirow{4}{*}{\rotatebox[origin=c]{45}{Qw-A}} 
 & \xmark & \xmark & 42.43${\pm2.84}$ & 43.93${\pm6.14}$ & 25.97${\pm4.44}$ & 20.43${\pm4.68}$ & 25.28${\pm1.39}$ & 22.71${\pm4.78}$ \\
 & \cmark & \xmark & 55.94${\pm7.08}$ & 53.94${\pm3.57}$ & 19.13${\pm2.81}$ & 27.33${\pm2.35}$ & 26.17${\pm1.65}$ & 30.19${\pm3.05}$ \\
 & \xmark & \cmark & 33.02${\pm0.00}$ & 33.02${\pm0.00}$ & 21.38${\pm1.65}$ & 22.34${\pm3.80}$ & 19.86${\pm4.40}$ & 18.99${\pm4.36}$ \\
 & \cmark & \cmark & 33.62${\pm1.21}$ & 33.43${\pm1.33}$ & 21.97${\pm2.91}$ & 25.80${\pm3.49}$ & 27.85${\pm1.76}$ & 28.58${\pm1.83}$ \\
\cmidrule(lr){1-9}
\multirow{4}{*}{\rotatebox[origin=c]{45}{MiniCPM}} 
 & \xmark & \xmark & 56.59${\pm9.37}$ & 38.51${\pm3.04}$ & 19.27${\pm0.46}$ & 19.40${\pm0.34}$ & 27.68${\pm2.74}$ & 25.58${\pm0.91}$ \\
 & \cmark & \xmark & 38.19${\pm2.94}$ & 43.63${\pm5.74}$ & 18.01${\pm0.00}$ & 20.42${\pm1.80}$ & 25.76${\pm1.30}$ & 26.66${\pm1.86}$ \\
 & \xmark & \cmark & 33.62${\pm1.21}$ & 35.98${\pm2.64}$ & 20.88${\pm0.67}$ & 20.07${\pm0.30}$ & 26.49${\pm2.33}$ & 23.97${\pm1.71}$ \\
 & \cmark & \cmark & 52.41${\pm3.74}$ & 47.48${\pm2.42}$ & 18.11${\pm1.09}$ & 18.92${\pm1.86}$ & 22.55${\pm1.01}$ & 26.29${\pm3.37}$ \\
\cmidrule(lr){1-9}
\multirow{4}{*}{\rotatebox[origin=c]{45}{Phi-4}} 
 & \xmark & \xmark & 45.31${\pm3.14}$ & 46.16${\pm1.80}$ & \underline{29.60${\pm2.37}$} & 30.19${\pm4.23}$ & \underline{31.03${\pm1.41}$} & 28.36${\pm2.42}$ \\
 & \cmark & \xmark & 47.62${\pm3.61}$ & 49.96${\pm5.60}$ & 24.36${\pm3.26}$ & 29.42${\pm3.03}$ & 29.83${\pm2.30}$ & 29.61${\pm1.71}$ \\
 & \xmark & \cmark & 37.38${\pm3.41}$ & 40.49${\pm4.29}$ & 19.50${\pm0.00}$ & 19.89${\pm0.78}$ & 23.85${\pm0.64}$ & 24.96${\pm2.01}$ \\
 & \cmark & \cmark & 44.20${\pm3.21}$ & 41.74${\pm3.85}$ & 19.04${\pm1.79}$ & 24.88${\pm5.10}$ & 26.07${\pm1.16}$ & 29.58${\pm1.80}$ \\
\cmidrule(lr){1-9}
\multirow{4}{*}{\rotatebox[origin=c]{45}{Qw-O-3B}} 
 & \xmark & \xmark & 39.61${\pm3.24}$ & 44.05${\pm9.71}$ & 24.42${\pm1.79}$ & 23.75${\pm3.31}$ & 28.07${\pm2.45}$ & 21.88${\pm5.80}$ \\
 & \cmark & \xmark & 60.01${\pm7.60}$ & 49.99${\pm8.26}$ & 22.22${\pm2.62}$ & 26.86${\pm2.86}$ & 29.58${\pm1.13}$ & 29.72${\pm0.41}$ \\
 & \xmark & \cmark & 48.25${\pm7.60}$ & 37.61${\pm7.12}$ & 20.23${\pm1.39}$ & 19.90${\pm0.84}$ & 26.64${\pm2.12}$ & 21.69${\pm6.00}$ \\
 & \cmark & \cmark & 60.26${\pm3.53}$ & 53.02${\pm7.02}$ & 22.96${\pm2.50}$ & 26.41${\pm1.86}$ & 29.03${\pm1.87}$ & 27.63${\pm3.38}$ \\
\cmidrule(lr){1-9}
\multirow{4}{*}{\rotatebox[origin=c]{45}{Qw-O-7B}} 
 & \xmark & \xmark & 38.52${\pm9.56}$ & 34.35${\pm2.67}$ & 24.31${\pm2.11}$ & 22.81${\pm3.99}$ & 28.52${\pm0.79}$ & 25.59${\pm3.85}$ \\
 & \cmark & \xmark & \underline{61.43${\pm5.55}$} & \underline{56.00${\pm5.47}$} & 26.22${\pm1.74}$ & 29.24${\pm4.96}$ & 29.93${\pm1.77}$ & 28.99${\pm1.71}$ \\
 & \xmark & \cmark & 37.80${\pm8.13}$ & 34.23${\pm1.48}$ & 27.47${\pm3.18}$ & 23.66${\pm4.76}$ & 28.70${\pm0.55}$ & 27.28${\pm2.19}$ \\
 & \cmark & \cmark & \textbf{\boldmath 66.13${\pm3.78}$} & \textbf{\boldmath 61.10${\pm6.79}$} & 27.77${\pm2.42}$ & \underline{31.19${\pm4.55}$} & 26.16${\pm1.90}$ & 27.74${\pm0.95}$ \\
\bottomrule
\end{tabular}
\caption{
The Avg@1 and Maj@5 (\%) of LLMs under four prompt types on benchmark datasets. The Avg@1 reflects single-shot performance based on a single sampled response per input, and the Maj@5 represents majority-vote accuracy aggregated over five sampled responses per input.
}

\label{tab:bench}
\end{table*}

\begin{figure*}[t]
    \centering
    \includegraphics[width=0.95\linewidth]{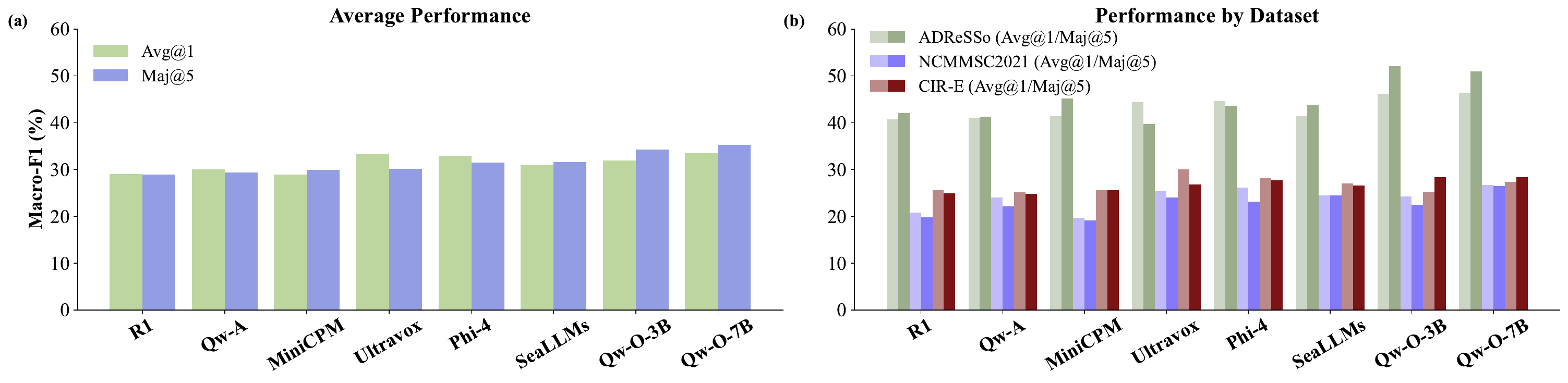}
    \caption{Comprehensive evaluation of LLMs on Avg@1 and Maj@5 metrics. (a) shows the overall performance of different LLMs averaged across three datasets. (b) shows detailed performance of each LLM on individual datasets.}

    \label{fig:table_bar}
\end{figure*}

\subsection{Large Language Model Performance}

To ensure fairness and consistency, we systematically evaluate several mainstream MLLMs on the proposed task within a unified framework. The MLLMs include R1-AQA~\cite{li2025reinforcement}, Ultravox-v0.5-llama-3.1-8b, SeaLLMs-Audio-7B~\cite{SeaLLMs-Audio}, Qwen2-Audio-7B-Instruct~\cite{Qwen2-Audio}, MiniCPM-o-2.6~\cite{yao2024minicpm}, Phi-4-Multimodal-Instruct~\cite{abouelenin2025phi}, as well as Qwen2.5-Omni-3B and Qwen2.5-Omni-7B~\cite{xu2025qwen2}. All MLLMs receive identical multimodal inputs consisting of the subject’s raw audio along with its corresponding ASR transcript.

\subsubsection{LLMs Baselines}

The prompt design follows a zero-shot paradigm to assess the model’s inherent task understanding and generalization. To further improve model performance, we explored two prompt enhancement strategies. The first is CoT prompt, which introduces instructions such as \textit{``please reason step by step''} to guide the model to perform explicit multi-step logical deduction to enhance its reasoning depth and logical consistency when handling complex tasks. The second is EXP injection, which incorporates key features that clinicians pay attention to during evaluation (such as language fluency, emotional expression, vocabulary selection, \textit{etc.}) into the prompt to simulate a professional evaluation framework, aiming to improve the clinical interpretability and professionalism of the model's judgment.

We evaluate models using two Macro-F1-based metrics: Avg@1, which considers only one prediction to reflect single-shot performance, and Maj@5, which aggregates predictions via majority voting over five samples per input.  
As shown in Table~\ref{tab:bench}, the comprehensive results provide a systematic comparison of multiple mainstream large language models on zero-shot cognitive screening across several benchmark datasets. The results demonstrate that CoT prompting significantly improves model performance, especially on the ADReSSo dataset. Most models show substantial performance gains when applying CoT, underscoring the effectiveness of this strategy in cognitive reasoning tasks.  
In contrast, the EXP prompt alone offers limited improvements and sometimes even degrades performance, possibly due to the introduction of distracting or misleading information. The combined CoT and EXP prompt results are mixed; the success of EXP prompts depends heavily on careful and precise prompt design, otherwise they may introduce noise and weaken model performance.

Figure~\ref{fig:table_bar} illustrates that the Maj@5 strategy is partially effective for some models but does not universally improve performance; thus, ensemble approaches should consider each model’s stability and diversity. Among all models, the Qw-O series consistently achieves the best results on both metrics, particularly excelling in Maj@5, demonstrating strong prediction consistency and generalization ability. At the dataset level, models perform better on the English binary classification dataset ADReSSo than on the Mandarin ternary classification datasets NCMMSC2021 and CIR-E, showing a trend opposite to that observed in SSMs. Performance on NCMMSC2021 is notably lower than on the other two datasets, likely due to its diverse task types causing prompt-context inconsistency, which hinders effective prompt-based reasoning.

\subsubsection{Fine-tuning}
\label{Fine-tuning}

\begin{table}[t]
\centering
\small
\begin{tabular}{c|c|ccc}
\toprule
\multirow{2}{*}{\textbf{Train}}    & \multirow{2}{*}{\textbf{Rate}}  & \multicolumn{3}{c}{\textbf{Test}}\\

 & &  \textbf{ADReSSo} & \textbf{NCMMSC} & \textbf{CIR-E} \\

\midrule
   \multirow{4}{*}{\rotatebox[origin=c]{90}{ADReSSo}}   
     &  \makebox[2em][r]{0\,\%}   & \cellcolor{gray!20} 55.94$_{\pm7.08}$  & 19.13$_{\pm2.81}$  & 26.17$_{\pm1.65}$  \\
     & \makebox[2em][r]{20\%}  & \cellcolor{gray!20} 62.52$_{\pm3.94}$  & 20.15$_{\pm2.74}$ & 28.84$_{\pm2.26}$ \\
    & \makebox[2em][r]{50\%}  & \cellcolor{gray!20} \underline{70.77$_{\pm1.23}$}  & \underline{20.41$_{\pm1.93}$}  & \textbf{\boldmath 30.93$_{\pm2.47}$} \\
     & \makebox[2em][r]{100\%}  & \cellcolor{gray!20} \textbf{\boldmath 74.69$_{\pm2.15}$}   &  \textbf{\boldmath 21.30$_{\pm1.76}$} &  \underline{28.49$_{\pm2.42}$} \\
\midrule

\multirow{4}{*}{\rotatebox[origin=c]{90}{NCMMSC}}
   & \makebox[2em][r]{0\,\%}  & 55.94$_{\pm7.08}$  & \cellcolor{gray!20} 19.13$_{\pm2.81}$   & 26.17$_{\pm1.65}$ \\

    & \makebox[2em][r]{20\%}   & \underline{63.89$_{\pm4.52}$}  & \cellcolor{gray!20} 57.62$_{\pm3.89}$ & 36.46$_{\pm2.61}$ \\

   & \makebox[2em][r]{50\%}  & \textbf{\boldmath 64.16$_{\pm5.60}$}  & \cellcolor{gray!20} \underline{68.09$_{\pm1.81}$}  & \underline{43.70$_{\pm1.94}$} \\

    & \makebox[2em][r]{100\%}   & 61.87$_{\pm3.15}$ & \cellcolor{gray!20} \textbf{\boldmath 71.36$_{\pm1.58}$} & \textbf{\boldmath 50.98$_{\pm0.97}$} \\

\bottomrule
\end{tabular}
\caption{
Performance of Qw-A w/ LoRA using 20\%, 50\%, and 100\% of training data, evaluated by Maj@5.
}
\label{tab:LoRA}
\end{table}

To reduce the computational cost typically associated with full-parameter fine-tuning of LLMs, we adopt a parameter-efficient strategy based on LoRA, which introduces lightweight trainable modules instead of updating the entire model.

To improve the robustness of MLLM-based classification, we move beyond direct label prediction, which often suffers from instability and low consistency. Instead, we observe that predictions accompanied by CoT reasoning tend to be more stable and interpretable. Based on this insight, we construct a high-quality fine-tuning corpus by generating CoT-style examples through a reverse prompting strategy.
Specifically, we design a prompt that takes as input the subject's audio and ASR transcription, along with the ground-truth label. The model is then asked to assess the subject's speech and language characteristics and explain the underlying condition indicated by the label. This process yields detailed CoT reasoning traces aligned with the true diagnosis, forming a set of supervision-ready training samples.

These CoT-augmented examples are used for instruction tuning with a parameter-efficient approach based on LoRA. As shown in Table~\ref{tab:LoRA}, we conduct experiments on the Qwen2-Audio model using 20\%, 50\%, and 100\% of the generated data, demonstrating the effectiveness of this approach in leveraging model-generated reasoning for adaptation.

\subsubsection{Comparing the Performance of SSMs and LLMs}

\begin{figure}[t]
    \centering
    \includegraphics[width=0.98\linewidth]{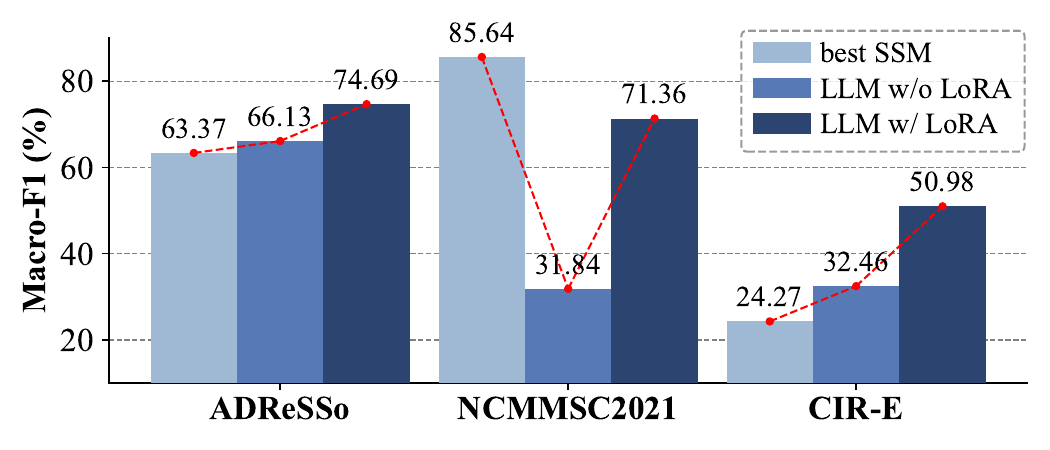}

    \caption{Performance comparison of LLM w/ and w/o LoRA against best SSM across three datasets.}

    \label{fig:Compaire_Performance}
\end{figure}

Figure~\ref{fig:Compaire_Performance} compares the performance of SSMs, LLMs w/o LoRA, and LLMs w/ LoRA across three datasets. Overall, LLMs w/o LoRA slightly outperform SSMs on ADReSSo and CIR-E, both of which feature standard picture description tasks. 
However, their performance drops significantly on NCMMSC2021, where SSMs demonstrate superior performance with clearer class boundaries. This can be partly attributed to the biased data distributions across classes in NCMMSC2021, allowing supervised SSMs to more easily exploit distribution-specific patterns. However, this advantage may not reflect a deeper understanding of cognitive impairment compared to LLMs.
In contrast, LLMs with LoRA consistently outperform basic LLMs across all datasets. Notably, on the OOD CIR-E test set, LoRA-enhanced LLMs show superior generalization ability compared to SSMs.

\subsubsection{Test-time Scaling}
\label{tts}

To assess the robustness and stability of model predictions during inference, we adopt a test-time scaling (TTS) strategy that performs majority voting over $K$ repeated outputs. This approach harnesses the natural variability in generative outputs to mitigate stochastic errors and better capture the model’s true confidence. 

\begin{figure}[t]
    \centering
    \includegraphics[width=0.98\linewidth]{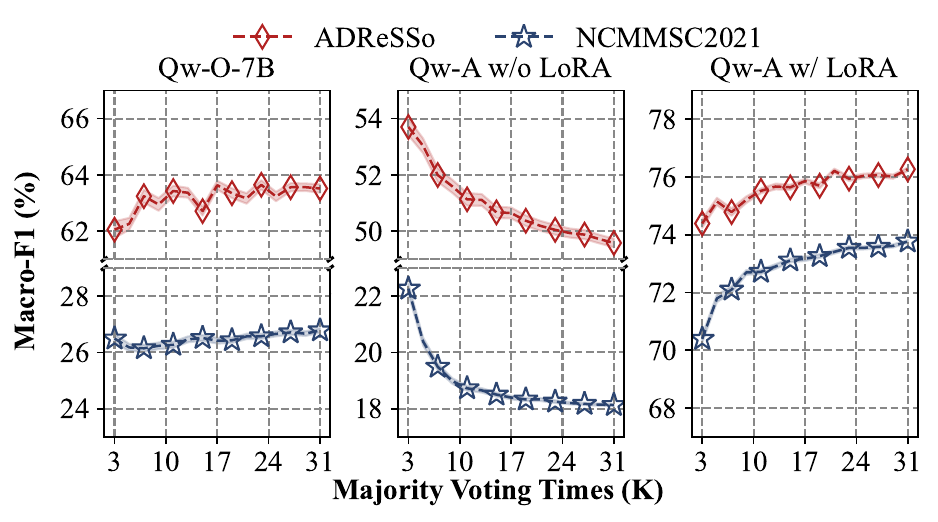}
    \caption{The Maj@K with different majority voting times K for three models under TTS.}

    \label{fig:test_time_scaling}
\end{figure}

Figure~\ref{fig:test_time_scaling} presents the test results, where we evaluate the best-performing Qw-O-7B model alongside Qw-A both with and without LoRA. Notably, the Qw-O-7B demonstrates strong compatibility with TTS, indicating more robust and consistent predictions for cognitive assessment tasks. In contrast, Qw-A w/o LoRA does not benefit from TTS; its performance even declines as $K$ increases, suggesting instability under stochastic sampling. However, Qw-A w/ LoRA exhibits a clear ability to detect cognitive impairment in speech-based cognitive assessment, although this capability is not fully activated during single-shot inference. TTS helps recover this potential by reducing output variability and mitigating performance degradation caused by suboptimal samples.


\subsubsection{Case Study}

To gain deeper insights into the decision-making and limitations of LLMs, we analyze two representative failure cases. By reviewing their CoT outputs, we aim to uncover the underlying causes of these errors. Detailed reasoning traces are provided in the Appendix.

\textbf{Subject 1:} This case comes from the ADReSSo. The subject's cognitive function is normal, with an MMSE score of 30/30. Although the LLMs predicted the subject as cognitively impaired, this constitutes a clear false positive.
The transcript reflects logical event progression (\textit{e.g.}, \textit{``the stool is falling over,'' ``he’s grabbing the cookie''}), multiple agent-action-object constructions, and an overall coherent narrative describing the Cookie Theft picture. The minor sentence-final interruption (\textit{``and the child is…''}) appears to stem from natural hesitancy or time constraints, rather than any underlying cognitive dysfunction.
The model’s rationale misinterprets this benign disfluency as indicative of impairment, demonstrating an over-sensitivity to surface-level pauses and an underestimation of global coherence and content structure. This case highlights a key limitation of LLM-based evaluations: without multimodal grounding or better calibration to normative variations in spontaneous speech—especially among older adults—such models risk overpathologizing normal behavior.

\textbf{Subject 2:} This subject comes from the CIR-E. He is a 75-year-old male with secondary education, clinically diagnosed with AD. He scored 25/30 on the MMSE and 18/30 on the MoCA. However, the report repeatedly classified him as a HC. The model’s rationale cited fluent speech and coherent sentence structure in his ASR transcript (\textit{e.g.}, ``this stool is about to fall, it's dangerous...'') as evidence of preserved cognition. In reality, the subject's description was overly brief, repetitive, and lacked contextual richness, which are hallmarks of semantic impoverishment often seen in AD. This misclassification highlights a key limitation of current LLM-based assessments: they over-rely on surface-level fluency while failing to capture deeper content deficits. Without grounding in clinical context and semantic expectations, such models risk under-pathologizing impaired individuals who retain superficial linguistic fluency.

\section{Discussion}

Despite recent advances in MLLMs, their application to speech-based cognitive assessment remains limited by several key challenges. First, MLLMs struggle to effectively capture salient acoustic features that are critical for distinguishing cognitive status. Cognitive decline is strongly correlated with aging, and vocal characteristics can vary significantly across age groups. In datasets such as NCMMSC, smaller models have occasionally outperformed MLLMs. A closer analysis reveals that this is partially due to the distinct age distributions between cognitively impaired and healthy participants. Speech from healthy individuals tends to be louder, more assertive, and more concise, while cognitively impaired speech is often softer and more hesitant. These nuances are not always captured by general-purpose MLLMs, which are not explicitly tuned to such demographic or clinical variations.

Second, current MLLMs show excessive sensitivity to disfluencies and repetitions—speech phenomena that are both common in patients with cognitive impairment and diagnostically relevant. This sensitivity is amplified by two compounding factors: the limited size of training datasets, which hinders the model’s ability to generalize across speaker variability, and the use of single-task in multimodal inputs, which constrains the model’s ability to objectively assess the cognitive status.

To address these limitations, a promising future direction would be to explore a multimodal optimization framework from two complementary perspectives: 1) Introducing baseline patient information ((\textit{e.g.}, age, gender, education, health records) into the prompt to provide personalized context for model inference; 2) Extracting acoustic features related to speech fluency and vocal intensity into the input stream, allowing the model to better differentiate between pathological and non-pathological variations in expressive patterns; 3) Employing a conditional data generation pipeline that synthesizes diverse speech-text samples based on patient baseline profiles; 4) Adopting a staged task design by initializing low-complexity voice interaction tasks for setting an individual's expressive baseline. This hybrid framework will enable MLLMs to calibrate their outputs and step-by-step reasoning.

\section{Conclusion}

In this work, we addressed the task of multilingual and cross-site cognitive impairment assessment from speech, a critical challenge in building practical AI-assisted diagnostic tools. We proposed \textbf{CogBench}, the first benchmark to systematically evaluate the generalization capabilities of both traditional neural models and LLMs across languages and clinical sites.

\bibliography{aaai2026}

\clearpage

\section{Appendix}

This Appendix is structured as follows. Appendix~A provides details about the datasets used, with a particular focus on the population distribution in the newly proposed CIR-E dataset. Appendix~B outlines the experimental settings, including both hardware and software environments, as well as the evaluation metrics. Appendix~C describes the experimental setup for SSMs. Appendix~D presents the experimental setup for LLMs, along with additional results and analysis.

\section{A. Datasets}

The INTERSPEECH2021 \textbf{ADReSSo} dataset is an extended version of the INTERSPEECH2020 ADReSS dataset, consisting of English speech recordings based on the ``\textit{Cookie Theft}'' picture description task for Alzheimer's disease detection. The \textbf{NCMMSC2021-AD} dataset, provided by the NCMMSC2021 Challenge, contains Mandarin speech recordings involving picture description, fluency tests, and self-introductions, aimed at detecting Alzheimer's disease and mild cognitive impairment.

Furthermore, we collected an additional test set, \textbf{CIR-E}, to evaluate our model in real-world scenarios. The CIR-E dataset was collected from elderly communities located in Jiangsu Province, China. It consists of speech samples recorded during picture description tasks, conducted by community-dwelling elderly participants under the guidance of clinical professionals. This dataset reflects natural and spontaneous speech in practical conditions and serves as a valuable resource for assessing model robustness. The detailed statistics of all datasets are summarized in Table~\ref{tab:dataset}.
All speech samples in the CIR-E dataset were collected via a standardized picture description task, with each recording lasting no more than one minute. Medical staff first conducted a preliminary screening among elderly individuals residing in community centers who met the target age criteria. We applied the following exclusion criteria:

\begin{enumerate}[label=\arabic*), itemsep=2pt, topsep=2pt, parsep=0pt]
    \item \textbf{Neurological Disorders}: Individuals with a history or current diagnosis of neurological diseases that may impair cognition, such as cerebrovascular disease, traumatic brain injury, epilepsy, or Parkinson’s disease.
    
    \item \textbf{Psychiatric Disorders}: Individuals with severe psychiatric conditions (e.g., major depression, schizophrenia) that were unstable or poorly managed.
    
    \item \textbf{Severe Systemic Diseases}: Individuals with serious liver or kidney dysfunction, or multi-organ failure.
    
    \item \textbf{Polypharmacy}: Individuals undergoing complex medication regimens that could significantly impact cognitive function.
    
    \item \textbf{Screening and Compliance Issues}: Individuals unable to complete scale-based assessments or with communication barriers that hindered study participation.
\end{enumerate}

\begin{table}[t]
\centering
\small
\begin{tabular}{cccccc}
\toprule
  & \textbf{Split} & \textbf{Cls} & \textbf{Subj. ($n$)} & \textbf{Samp. ($n$)} & \textbf{Dur. ($s$)} \\
\midrule
\multirow{4}{*}{\rotatebox[origin=c]{90}{ADReSSo}} 
& \multirow{2}{*}{Train} 
         & Non-AD & 79 & 79 & $22 \sim 162$ \\
&        & AD     & 87 & 87 & $19 \sim 226$ \\

\cmidrule{2-6}
& \multirow{2}{*}{Test} 
       & Non-AD & 36 & 36 & $22 \sim 134$ \\
&        & AD     & 35 & 35 & $20 \sim 136$ \\

\midrule
\multirow{6}{*}{\rotatebox[origin=c]{90}{NCMMSC2021}} 
& \multirow{3}{*}{Train} 
 & HC   & 44 & 108 & $28 \sim 60$ \\
 & & MCI  & 53 & 93  & $28 \sim 60$ \\
 & & AD   & 26 & 79  & $28 \sim 60$ \\
\cmidrule{2-6}
& \multirow{3}{*}{Test} 
 & AD   & 10 & 35  & $50 \sim 60$ \\
 & & MCI  & 23 & 39  & $44 \sim 60$ \\
 & & HC   & 20 & 45  & $47 \sim 60$ \\

\midrule
\multirow{3}{*}{\rotatebox[origin=c]{90}{CIR-E}} 
& \multirow{3}{*}{Test} 
  & HC   & 11 & 33 & $10 \sim 60$ \\
 & & MCI  & 27 & 74  & $9 \sim 60$ \\
 & & AD   & 16 & 46  & $17 \sim 60$ \\
\bottomrule
\end{tabular}
\caption{
  Summary of the datasets used in our benchmark: ADReSSo, NCMMSC2021-AD, and CIR-E. 
  For each subset, we report the number of diagnosis classes (\textbf{Cls}), subjects (\textbf{Subj.}), speech samples (\textbf{Samp.}), and the duration range in seconds (\textbf{Dur.}). 
  Diagnosis categories include HC, MCI, and AD. 
  When applicable, data is split into training and testing sets. 
  Note that CIR-E is used exclusively for testing to evaluate generalization performance.
}

\label{tab:dataset}
\end{table}

\begin{figure}[t]
    \centering
    \includegraphics[width=0.98\linewidth]{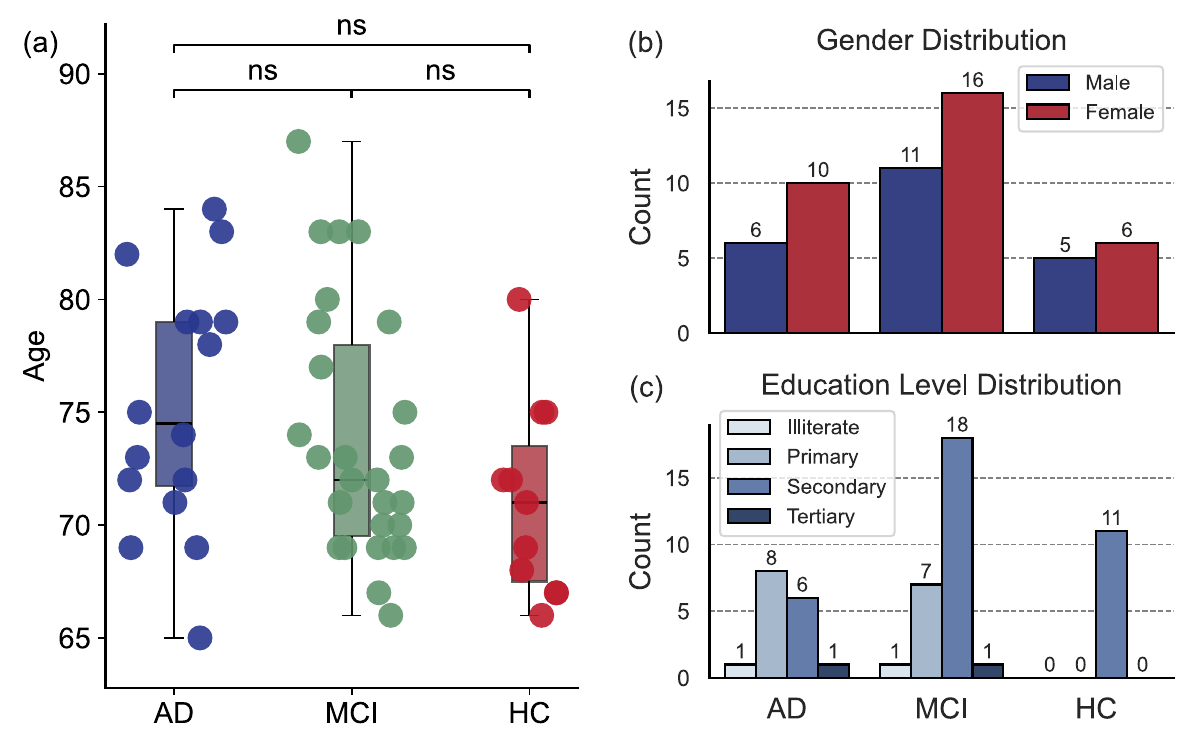}
    \caption{
    Demographic distribution of participants in the CIR-E dataset.  
    (a) Age distribution across diagnostic categories; no significant differences were observed between groups.  
    (b) Gender distribution across categories.  
    (c) Distribution of education levels, including illiterate, primary school, secondary school, and tertiary education (college or above).
    This figure illustrates the balance in key demographic variables, helping to control for potential confounding effects.
    }
    \label{fig:CIR-E}
\end{figure}

For individuals who passed the initial screening, the study protocol was fully explained, and informed consent was obtained. Comprehensive clinical assessments were subsequently conducted by experienced neurologists, incorporating standardized cognitive evaluations, including Mini-Mental State Examination (MMSE), the Montreal Cognitive Assessment (MoCA), and other comprehensive neuropsychological and functional assessments.

Participant group allocation was determined based on cognitive performance, medical history, and physical examination. The control group consisted of cognitively normal individuals, without subjective memory complaints or any history of major neurological, psychiatric, or metabolic disorders.
Figure~\ref{fig:CIR-E} illustrates the demographic composition of the diagnostic groups. There were no statistically significant differences in age (ANOVA $p = 0.148$) or gender (Chi-square $= 0.171$, $p = 0.918$) across diagnostic groups, which may help mitigate potential demographic confounding effects.

\section{B. Experimental Setup}

\subsection{Hardware and Environment}

All experiments were conducted on a high-performance computing cluster equipped with 16 NVIDIA RTX 3090 GPUs, running CUDA 12.4. The software environment included PyTorch 2.6, along with supporting libraries such as \textit{torchaudio}, \textit{librosa}, \textit{transformers}, and \textit{vllm}.

\subsection{Evaluation Metrics}

\textbf{Accuracy} measures the overall correctness of the model by calculating the proportion of correctly predicted samples among all samples:
\begin{equation}
\text{Accuracy} = \frac{\sum_{i=1}^{C} TP_i}{N},
\end{equation}
where \(C\) is the number of classes, \(TP_i\) denotes the number of true positives for class \(i\), and \(N\) is the total number of samples.

\textbf{Precision} for each class measures the proportion of correctly predicted positive samples out of all samples predicted as positive:
\begin{equation}
\text{Precision}_i = \frac{TP_i}{TP_i + FP_i}.
\end{equation}

\textbf{Recall} for each class measures the proportion of correctly predicted positive samples out of all actual positive samples:
\begin{equation}
\text{Recall}_i = \frac{TP_i}{TP_i + FN_i}.
\end{equation}

\textbf{Macro-F1} is commonly used in multi-class classification, especially when classes are imbalanced. It calculates the F1-score for each class independently and averages them equally:
\begin{equation}
\text{Macro-F1} = \frac{1}{C} \sum_{i=1}^C \text{F1}_i,
\end{equation}
where the class-wise F1-score is the harmonic mean of precision and recall:
\begin{equation}
\text{F1}_i = \frac{2 \cdot \text{Precision}_i \cdot \text{Recall}_i}{\text{Precision}_i + \text{Recall}_i}.
\end{equation}

Here, \(TP_i\), \(FP_i\), and \(FN_i\) denote the true positives, false positives, and false negatives for class \(i\), respectively. Macro-F1 treats all classes equally, providing a balanced metric that reflects performance across categories.

\section{C. Small-Scale Models}

To establish a supervised learning baseline, we trained four SSMs: two time-domain models operating directly on raw audio waveforms—\textit{1D-ResNet} and \textit{LSTM}—and two frequency-domain models using mel-spectrogram features—\textit{ResNet18} and \textit{Transformer}.

Given the variability in recording lengths, all audio data were segmented using a 6-second sliding window with a 2-second stride. During inference, predictions from individual segments were aggregated via majority voting to produce the final subject-level decision.

Training was performed using the Adam optimizer with a cosine learning rate decay schedule. To identify optimal hyperparameters, we conducted a grid search over learning rates $\{0.005,\ 0.0001,\ 0.0003\}$ and batch sizes $\{32,\ 64,\ 128,\ 512\}$. To ensure robustness and mitigate randomness, each configuration was evaluated using five different random seeds, and the final performance was reported as the average across these runs.
Table~\ref{tab:gridsearch} summarizes the best hyperparameter configurations identified for each model and dataset.

\begin{table}[h]
\centering
\small
\begin{tabular}{clcc}
\toprule
\textbf{Dataset} & \textbf{Model} & \textbf{Learning Rate} & \textbf{Batch Size} \\

\midrule
\multirow{4}{*}{\rotatebox[origin=c]{45}{ADReSSo}} 
         & 1D-ResNet     & 0.0003 &  64  \\
         & LSTM          & 0.005 &  512  \\
         & ResNet18      & 0.0001 & 512   \\
         & Transformer   & 0.0001 &  512  \\
\midrule
\multirow{4}{*}{\rotatebox[origin=c]{45}{NCMMSC}} 
  & 1D-ResNet     & 0.0003 &  32  \\
  & LSTM          & 0.005 & 128  \\
  & ResNet18      & 0.0001 &  512  \\
  & Transformer   & 0.0003 &  64 \\
\bottomrule
\end{tabular}
\caption{Grid search for optimal parameters results}
\label{tab:gridsearch}
\end{table}


\begin{table*}[t]
\centering
\small
\begin{tabular}{c|l|c|l}
\toprule
\textbf{Type} & \textbf{Model} & \textbf{Size} & \textbf{Link} \\
\midrule
\multirow{4}{*}{Audio} 
     & R1-AQA  & 7B & \url{https://huggingface.co/mispeech/r1-aqa} \\
     & Ultravox-v0.5-llama-3.1-8b &  8B & \url{https://huggingface.co/fixie-ai/ultravox-v0_5-llama-3_1-8b} \\
     & SeaLLMs-Audio-7B & 7B & \url{https://huggingface.co/SeaLLMs/SeaLLMs-Audio-7B} \\
     & Qwen2-Audio-7B-Instruct & 7B & \url{https://huggingface.co/Qwen/Qwen2-Audio-7B-Instruct} \\
\midrule
\multirow{4}{*}{Omni} 
      & MiniCPM-o-2.6 &  8B & \url{https://huggingface.co/openbmb/MiniCPM-o-2_6} \\
    & Phi-4-Multimodal-Instruct  & 5B & \url{https://huggingface.co/microsoft/Phi-4-multimodal-instruct} \\
      & Qwen2.5-Omni-3B  & 3B & \url{https://huggingface.co/Qwen/Qwen2.5-Omni-3B} \\
      & Qwen2.5-Omni-7B   & 7B & \url{https://huggingface.co/Qwen/Qwen2.5-Omni-7B} \\
\bottomrule
\end{tabular}
\caption{Model cards for LLMs.}
\label{tab:model_names}
\end{table*}

\section{D. Large Language Models}


We systematically evaluate the performance of several mainstream LLMs on the proposed task, including R1-AQA, Ultravox-v0.5-llama-3.1-8b, SeaLLMs-Audio-7B, Qwen2-Audio-7B-Instruct, MiniCPM-o-2.6, Phi-4-Multimodal-Instruct, as well as Qwen2.5-Omni-3B and Qwen2.5-Omni-7B. Table~\ref{tab:model_names} presents the basic information of these models, including the model type, parameter size, and official homepage link for further reference.

\begin{table}[h]
\centering
\small
\begin{tabular}{l|c}
\toprule
\textbf{Parameter} & \textbf{Value} \\
\midrule
\textit{dtype}              & bf16 \\
\textit{n\_sample}          & 5 \\
\textit{temperature}        & 0.7 \\
\textit{top\_p}             & 1 \\
\textit{top\_k}             & -1 \\
\textit{max\_model\_len}    & 8192 \\
\textit{max\_tokens}        & 1024 \\
\textit{max\_num\_seqs}     & 1 \\
\textit{tp\_size}           & 2 \\
\bottomrule
\end{tabular}
\caption{LLM Inference Parameter Settings}
\label{tab:llm_params}
\end{table}

\subsection{Implementation Details}

For LLM-based inference, we standardized the decoding parameters across all models to ensure fair comparison. Table~\ref{tab:llm_params} lists the key hyperparameters used during inference.

\subsection{Inference Prompt Design}

During the inference phase, we designed two types of prompt templates for LLMs: an English version for binary classification and a Chinese version for three-class classification. While differing in language and label granularity, both templates share the same structural format.

We embed the subject’s raw audio and ASR-transcribed text into the prompt template. The specific settings are as follows:

\begin{itemize}
    \item \textbf{English Binary Prompt:} used for English datasets such as \textit{ADReSSo}, where the task is to classify the transcript into either AD or Non-AD.
    \item \textbf{Chinese Ternary Prompt:} used for Chinese datasets such as \textit{NCMMSC2021} and \textit{CIR-E}, where the labels include HC, MCI, and AD.
\end{itemize}

The prompt design adopts a zero-shot paradigm to evaluate the large language models’ inherent understanding and generalization capabilities on the cognitive impairment classification task. To further enhance model performance, we investigated two prompt augmentation strategies, both individually and in combination.

\begin{figure}[t!]
    \centering
    \includegraphics[width=0.98\linewidth]{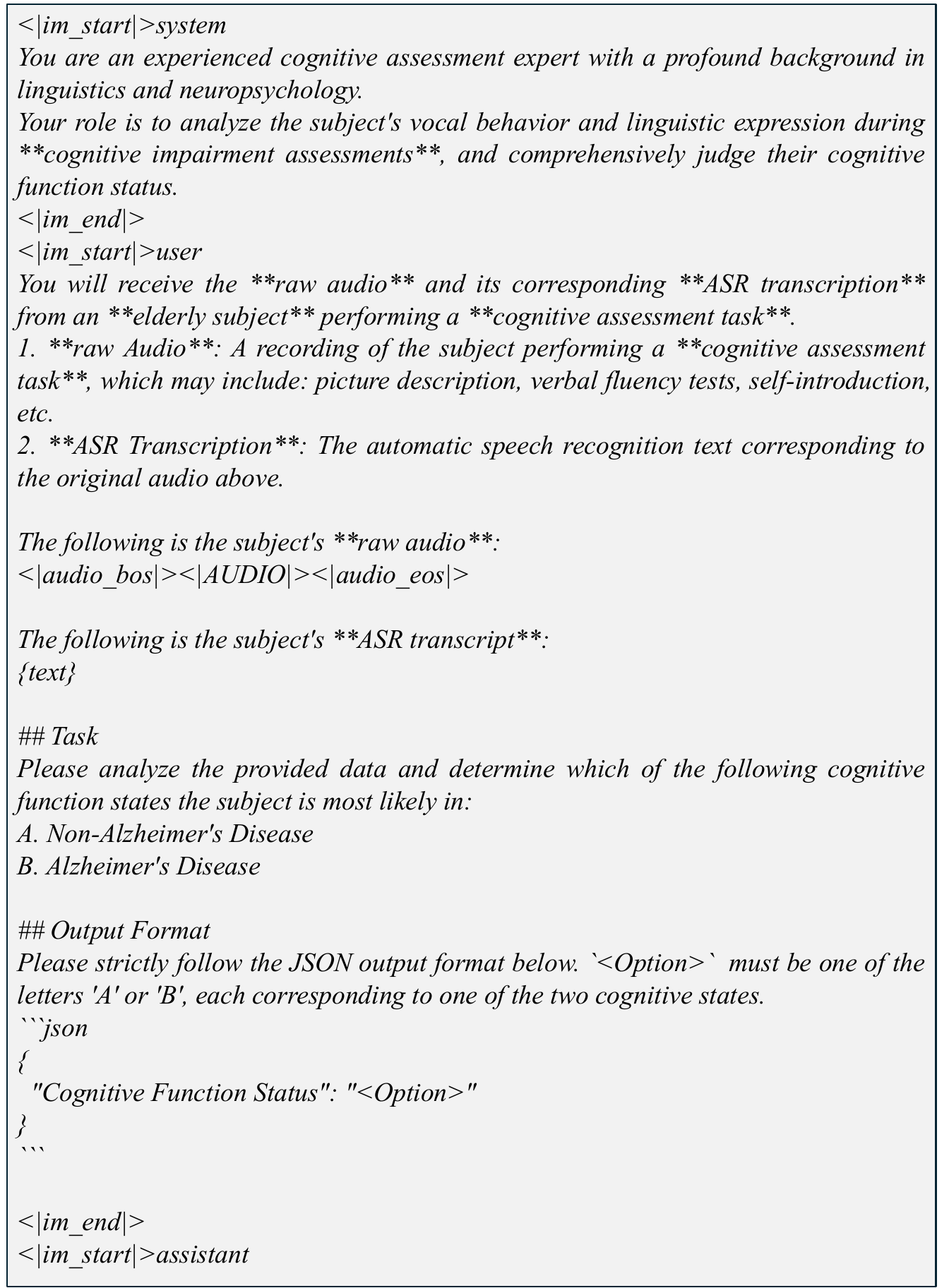}
    \caption{An example of the zero-shot prompt template, illustrated for the English binary classification task.}
    \label{fig:zero_shot}
\end{figure}

The first strategy is Chain-of-Thought (CoT) prompting, which includes explicit instructions like ``\textit{please reason step by step}'' to encourage the model to perform multi-step logical reasoning. This approach aims to deepen the model’s inference process and improve logical consistency, particularly when addressing complex or nuanced inputs.

The second strategy is Expert-knowledge  (EXP) injection, where clinically relevant indicators—such as language fluency, emotional expression, and vocabulary choice—that expert evaluators focus on during cognitive assessments are incorporated directly into the prompt. This simulates a professional clinical evaluation framework, with the goal of enhancing both the interpretability and clinical relevance of the model’s predictions.

Figure~\ref{fig:zero_shot} illustrates the zero-shot prompt template, which serves as the foundation for all subsequent prompt variants. Here, \texttt{\textless AUDIO\textgreater} and \texttt{\textless text\textgreater} represent the input transcript and audio, respectively.

For CoT prompting, we use the instruction:

\textit{"Please combine the subject's audio characteristics and language content, analyze and reason step-by-step, explain the rationale for your judgment, and ultimately output the categorical result."}

This encourages the model to engage in a step-by-step reasoning process rather than directly outputting the final decision.  
Moreover, the output format is extended to include a \textit{"Rationale"} field, which contains the model’s reasoning process before the final classification, thereby enhancing the interpretability of its judgment.

For EXP prompting, we introduce an instruction that simulates expert-level cognitive assessment by incorporating key clinical criteria and characteristic features of different cognitive states. The goal is to enhance the clinical interpretability and relevance of LLM outputs by guiding the model to evaluate both linguistic and vocal aspects, as detailed in the following prompt template.

\textit{Please refer to the following professional cognitive function assessment principles and key features of each condition to assist in judging the participant's cognitive state. Your judgment should comprehensively consider both the vocal and linguistic performance.}

\textit{Core Judgment Principles}

\textit{1.  Completeness and Accuracy of Language Content: Can the participant accurately and completely describe the core information of the image?}

\textit{2.  Fluency and Coherence of Expression: Is the participant's expression fluent and natural? Are their thoughts clear, organized, and logically connected?}

\textit{3.  Frequency and Salience of Impairment Features: Are there features in the participant's speech or language that are indicative of cognitive impairment? How frequently do these features occur? How severely do they affect communication efficiency and content accuracy?}

\textit{4.  Tip: When the participant exhibits features associated with multiple cognitive states, the classification should primarily be based on the dominant features that most significantly impact their overall cognitive function.}

\textit{[Key Differentiating Features for Each Status]}

\textit{A. Cognitively Normal:}

   \textit{- Overall language expression is fluent, relatively informative, and generally logical and clear.}
   
   \textit{- Speech is natural, articulation is clear; occasional normal hesitations or minor disfluencies may be present but do not significantly impact overall communication.}
   
\textit{B. Cognitively Impaired:}

   \textit{- Language expression may exhibit information omission, loose organization of content, reduced logical coherence, or difficulty in capturing the core information of the image.}
   
   \textit{- Speech may manifest as slowed or irregular speaking rate, increased pauses, slurred articulation, frequent hesitations or prolongations, or even fragmented expression.}
   
   \textit{- Overall communication efficiency and the clarity/accuracy of expression are affected to varying degrees.}

\subsection{More Results}

We conducted a detailed evaluation of the models across three benchmark datasets, including metrics such as Accuracy, Precision, Recall, and Macro-F1. Table~\ref{tab:adresso} presents the results of LLMs on the ADReSSo dataset, Table~\ref{tab:ncmmsc2021} shows the results on the NCMMSC2021 dataset, and Table~\ref{tab:cir-e} summarizes the results on the CIR-E dataset.

\begin{table*}[t]
\centering
\small
\begin{tabular}{l|cc|cccc}
\toprule
\multirow{2}{*}{\textbf{Model}} & \multicolumn{2}{c|}{\textbf{Prompt}} & \multicolumn{4}{c}{\textbf{ADReSSo}} \\
 & COT & EXP & Accuracy & Precision & Recall & Macro-F1 \\
\midrule
\multirow{4}{*}{R1-AQA} & \xmark & \xmark & 55.21${\pm2.42}$ & 65.86${\pm5.78}$ & 55.73${\pm2.32}$ & 47.61${\pm6.15}$ \\
 & \cmark & \xmark & 60.28${\pm2.07}$ & \textbf{\boldmath 73.31${\pm5.65}$} & 60.78${\pm2.05}$ & 54.59${\pm3.18}$ \\
 & \xmark & \cmark & 49.30${\pm0.00}$ & 24.65${\pm0.00}$ & 50.00${\pm0.00}$ & 33.02${\pm0.00}$ \\
 & \cmark & \cmark & 49.30${\pm0.00}$ & 24.65${\pm0.00}$ & 50.00${\pm0.00}$ & 33.02${\pm0.00}$ \\
\cmidrule(lr){1-7}
\multirow{4}{*}{Ultravox-v0.5-llama-3.1-8b} & \xmark & \xmark & 52.39${\pm0.56}$ & 67.94${\pm6.21}$ & 53.03${\pm0.57}$ & 40.62${\pm0.91}$ \\
 & \cmark & \xmark & 52.96${\pm2.76}$ & 66.11${\pm9.84}$ & 53.59${\pm2.73}$ & 41.56${\pm4.68}$ \\
 & \xmark & \cmark & 50.14${\pm1.12}$ & 44.86${\pm24.75}$ & 50.83${\pm1.11}$ & 34.80${\pm2.37}$ \\
 & \cmark & \cmark & 50.70${\pm1.54}$ & 52.45${\pm6.10}$ & 51.25${\pm1.48}$ & 41.65${\pm4.81}$ \\
\cmidrule(lr){1-7}
\multirow{4}{*}{SeaLLMs-Audio-7B} & \xmark & \xmark & 51.27${\pm2.61}$ & 36.59${\pm14.42}$ & 51.45${\pm2.39}$ & 40.32${\pm9.14}$ \\
 & \cmark & \xmark & 52.68${\pm2.90}$ & 52.74${\pm3.01}$ & 52.69${\pm2.94}$ & 52.56${\pm2.85}$ \\
 & \xmark & \cmark & 49.58${\pm0.56}$ & 34.72${\pm20.14}$ & 50.28${\pm0.56}$ & 33.62${\pm1.21}$ \\
 & \cmark & \cmark & 53.52${\pm6.49}$ & 50.23${\pm14.05}$ & 53.17${\pm6.64}$ & 48.49${\pm10.99}$ \\
\cmidrule(lr){1-7}
\multirow{4}{*}{Qwen2-Audio-7B-Instruct} & \xmark & \xmark & 53.24${\pm1.06}$ & 68.68${\pm5.69}$ & 53.86${\pm1.02}$ & 42.43${\pm2.84}$ \\
 & \cmark & \xmark & 61.13${\pm5.39}$ & \underline{71.37${\pm6.34}$} & 61.61${\pm5.34}$ & 55.94${\pm7.08}$ \\
 & \xmark & \cmark & 49.30${\pm0.00}$ & 24.65${\pm0.00}$ & 50.00${\pm0.00}$ & 33.02${\pm0.00}$ \\
 & \cmark & \cmark & 49.58${\pm0.56}$ & 34.72${\pm20.14}$ & 50.28${\pm0.56}$ & 33.62${\pm1.21}$ \\
\cmidrule(lr){1-7}
\multirow{4}{*}{MiniCPM-o.2.6} & \xmark & \xmark & 60.56${\pm6.17}$ & 64.19${\pm7.40}$ & 60.51${\pm6.01}$ & 56.59${\pm9.37}$ \\
 & \cmark & \xmark & 50.99${\pm1.64}$ & 56.08${\pm18.91}$ & 51.64${\pm1.64}$ & 38.19${\pm2.94}$ \\
 & \xmark & \cmark & 49.58${\pm0.56}$ & 34.72${\pm20.14}$ & 50.28${\pm0.56}$ & 33.62${\pm1.21}$ \\
 & \cmark & \cmark & 56.62${\pm2.73}$ & 61.94${\pm5.78}$ & 57.03${\pm2.72}$ & 52.41${\pm3.74}$ \\
\cmidrule(lr){1-7}
\multirow{4}{*}{Phi-4-multimodal-instruct} & \xmark & \xmark & 50.70${\pm1.78}$ & 51.68${\pm2.78}$ & 51.15${\pm1.75}$ & 45.31${\pm3.14}$ \\
 & \cmark & \xmark & 49.86${\pm3.84}$ & 49.79${\pm4.90}$ & 49.59${\pm3.81}$ & 47.62${\pm3.61}$ \\
 & \xmark & \cmark & 50.42${\pm1.05}$ & 52.60${\pm16.34}$ & 51.07${\pm1.02}$ & 37.38${\pm3.41}$ \\
 & \cmark & \cmark & 48.73${\pm2.90}$ & 48.92${\pm5.03}$ & 49.13${\pm2.93}$ & 44.20${\pm3.21}$ \\
\cmidrule(lr){1-7}
\multirow{4}{*}{Qwen2.5-Omni-3B} & \xmark & \xmark & 53.52${\pm1.54}$ & 66.09${\pm20.37}$ & 52.86${\pm1.57}$ & 39.61${\pm3.24}$ \\
 & \cmark & \xmark & 61.69${\pm6.75}$ & 63.21${\pm7.39}$ & 61.46${\pm6.81}$ & 60.01${\pm7.60}$ \\
 & \xmark & \cmark & 54.93${\pm3.33}$ & 62.78${\pm6.92}$ & 55.01${\pm3.13}$ & 48.25${\pm7.60}$ \\
 & \cmark & \cmark & 61.13${\pm3.63}$ & 62.71${\pm4.49}$ & 61.33${\pm3.66}$ & 60.26${\pm3.53}$ \\
\cmidrule(lr){1-7}
\multirow{4}{*}{Qwen2.5-Omni-7B} & \xmark & \xmark & 51.55${\pm3.84}$ & 42.06${\pm21.75}$ & 52.17${\pm3.68}$ & 38.52${\pm9.56}$ \\
 & \cmark & \xmark & \underline{63.38${\pm4.45}$} & 67.07${\pm4.45}$ & \underline{63.68${\pm4.41}$} & \underline{61.43${\pm5.55}$} \\
 & \xmark & \cmark & 51.55${\pm3.84}$ & 43.47${\pm23.15}$ & 52.21${\pm3.76}$ & 37.80${\pm8.13}$ \\
 & \cmark & \cmark & \textbf{\boldmath 66.48${\pm3.61}$} & 67.20${\pm3.92}$ & \textbf{\boldmath 66.50${\pm3.71}$} & \textbf{\boldmath 66.13${\pm3.78}$} \\
\bottomrule
\end{tabular}
\caption{Results of LLMs on the ADReSSo dataset (\%).}
\label{tab:adresso}
\end{table*}


\begin{table*}[t]
\centering
\small
\begin{tabular}{l|cc|cccc}
\toprule
\multirow{2}{*}{\textbf{Model}} & \multicolumn{2}{c|}{\textbf{Prompt}} & \multicolumn{4}{c}{\textbf{NCMMSC2021}} \\
 & COT & EXP & Accuracy & Precision & Recall & Macro-F1 \\
\midrule
\multirow{4}{*}{R1-AQA} & \xmark & \xmark & 33.17${\pm0.56}$ & 17.68${\pm13.43}$ & 33.48${\pm0.30}$ & 16.86${\pm0.70}$ \\
 & \cmark & \xmark & 29.91${\pm1.56}$ & 27.13${\pm13.36}$ & 30.04${\pm1.52}$ & 19.04${\pm1.79}$ \\
 & \xmark & \cmark & 32.94${\pm0.34}$ & 12.48${\pm2.79}$ & 33.52${\pm0.38}$ & 17.05${\pm0.81}$ \\
 & \cmark & \cmark & 32.77${\pm2.43}$ & 38.42${\pm10.19}$ & 32.67${\pm2.68}$ & 26.35${\pm3.22}$ \\
\cmidrule(lr){1-7}
\multirow{4}{*}{Ultravox-v0.5-llama-3.1-8b} & \xmark & \xmark & 34.45${\pm1.77}$ & \textbf{\boldmath 62.13${\pm8.25}$} & 35.14${\pm1.97}$ & 22.04${\pm3.17}$ \\
 & \cmark & \xmark & 33.27${\pm0.86}$ & 46.60${\pm19.15}$ & 33.85${\pm0.96}$ & 19.21${\pm2.13}$ \\
 & \xmark & \cmark & 37.14${\pm2.84}$ & \underline{46.94${\pm19.76}$} & \underline{38.44${\pm3.44}$} & 25.87${\pm5.79}$ \\
 & \cmark & \cmark & 36.30${\pm2.28}$ & 31.98${\pm8.39}$ & 38.00${\pm2.69}$ & 29.16${\pm2.41}$ \\
\cmidrule(lr){1-7}
\multirow{4}{*}{SeaLLMs-Audio-7B} & \xmark & \xmark & 32.10${\pm4.37}$ & 26.62${\pm6.43}$ & 32.27${\pm4.51}$ & 24.36${\pm5.66}$ \\
 & \cmark & \xmark & 35.63${\pm3.04}$ & 41.09${\pm11.65}$ & 33.80${\pm2.81}$ & \textbf{\boldmath 30.02${\pm2.81}$} \\
 & \xmark & \cmark & 32.27${\pm1.46}$ & 17.38${\pm13.49}$ & 33.48${\pm0.30}$ & 16.51${\pm0.91}$ \\
 & \cmark & \cmark & 32.94${\pm2.83}$ & 41.01${\pm6.10}$ & 33.10${\pm2.54}$ & 27.15${\pm3.85}$ \\
\cmidrule(lr){1-7}
\multirow{4}{*}{Qwen2-Audio-7B-Instruct} & \xmark & \xmark & 35.12${\pm1.11}$ & 23.90${\pm3.27}$ & 37.06${\pm1.74}$ & 25.97${\pm4.44}$ \\
 & \cmark & \xmark & 30.42${\pm1.94}$ & 36.55${\pm17.25}$ & 30.81${\pm1.99}$ & 19.13${\pm2.81}$ \\
 & \xmark & \cmark & 30.42${\pm1.23}$ & 24.83${\pm6.53}$ & 33.89${\pm1.51}$ & 21.38${\pm1.65}$ \\
 & \cmark & \cmark & 33.44${\pm1.63}$ & 36.45${\pm14.99}$ & 33.96${\pm1.82}$ & 21.97${\pm2.91}$ \\
\cmidrule(lr){1-7}
\multirow{4}{*}{MiniCPM-o.2.6} & \xmark & \xmark & 33.95${\pm1.01}$ & 40.32${\pm8.23}$ & 34.30${\pm1.02}$ & 19.27${\pm0.46}$ \\
 & \cmark & \xmark & 33.61${\pm0.00}$ & 44.35${\pm0.00}$ & 34.07${\pm0.00}$ & 18.01${\pm0.00}$ \\
 & \xmark & \cmark & 28.91${\pm2.23}$ & 36.55${\pm15.86}$ & 29.82${\pm2.17}$ & 20.88${\pm0.67}$ \\
 & \cmark & \cmark & 33.61${\pm0.53}$ & 44.37${\pm21.11}$ & 34.11${\pm0.54}$ & 18.11${\pm1.09}$ \\
\cmidrule(lr){1-7}
\multirow{4}{*}{Phi-4-multimodal-instruct} & \xmark & \xmark & \textbf{\boldmath 39.33${\pm1.45}$} & 40.53${\pm13.53}$ & \textbf{\boldmath 38.70${\pm1.45}$} & \underline{29.60${\pm2.37}$} \\
 & \cmark & \xmark & 36.47${\pm1.81}$ & 33.94${\pm4.34}$ & 36.34${\pm1.53}$ & 24.36${\pm3.26}$ \\
 & \xmark & \cmark & 34.45${\pm0.00}$ & 44.44${\pm0.00}$ & 34.81${\pm0.00}$ & 19.50${\pm0.00}$ \\
 & \cmark & \cmark & 33.78${\pm1.11}$ & 41.03${\pm16.47}$ & 34.30${\pm1.08}$ & 19.04${\pm1.79}$ \\
\cmidrule(lr){1-7}
\multirow{4}{*}{Qwen2.5-Omni-3B} & \xmark & \xmark & 33.44${\pm1.63}$ & 30.39${\pm1.72}$ & 33.18${\pm1.72}$ & 24.42${\pm1.79}$ \\
 & \cmark & \xmark & 30.75${\pm3.30}$ & 23.26${\pm4.00}$ & 30.32${\pm3.22}$ & 22.22${\pm2.62}$ \\
 & \xmark & \cmark & 31.59${\pm1.73}$ & 25.90${\pm3.12}$ & 31.59${\pm2.01}$ & 20.23${\pm1.39}$ \\
 & \cmark & \cmark & 31.59${\pm2.53}$ & 30.80${\pm14.38}$ & 31.20${\pm2.47}$ & 22.96${\pm2.50}$ \\
\cmidrule(lr){1-7}
\multirow{4}{*}{Qwen2.5-Omni-7B} & \xmark & \xmark & 35.63${\pm1.14}$ & 32.19${\pm2.03}$ & 35.42${\pm0.91}$ & 24.31${\pm2.11}$ \\
 & \cmark & \xmark & 35.12${\pm1.63}$ & 26.75${\pm1.72}$ & 34.38${\pm1.46}$ & 26.22${\pm1.74}$ \\
 & \xmark & \cmark & 36.97${\pm1.60}$ & 29.47${\pm3.25}$ & 35.72${\pm0.83}$ & 27.47${\pm3.18}$ \\
 & \cmark & \cmark & \underline{37.14${\pm1.35}$} & 36.23${\pm17.03}$ & 34.05${\pm1.47}$ & 27.77${\pm2.42}$ \\
\bottomrule
\end{tabular}
\caption{Results of LLMs on the NCMMSC2021 dataset (\%).}
\label{tab:ncmmsc2021}
\end{table*}


\begin{table*}[t]
\centering
\small
\begin{tabular}{l|cc|cccc}
\toprule
\multirow{2}{*}{\textbf{Model}} & \multicolumn{2}{c|}{\textbf{Prompt}} & \multicolumn{4}{c}{\textbf{CIR-E}} \\
 & COT & EXP & Accuracy & Precision & Recall & Macro-F1 \\
\midrule
\multirow{4}{*}{R1-AQA} & \xmark & \xmark & 48.56${\pm0.38}$ & 16.18${\pm0.13}$ & 33.33${\pm0.00}$ & 21.79${\pm0.12}$ \\
 & \cmark & \xmark & 44.44${\pm2.65}$ & 31.97${\pm13.94}$ & 33.93${\pm2.16}$ & 27.44${\pm2.42}$ \\
 & \xmark & \cmark & 48.37${\pm0.00}$ & 19.00${\pm5.75}$ & 33.50${\pm0.33}$ & 22.44${\pm1.42}$ \\
 & \cmark & \cmark & 43.14${\pm5.73}$ & 37.43${\pm14.33}$ & 35.27${\pm0.95}$ & 28.03${\pm2.81}$ \\
\cmidrule(lr){1-7}
\multirow{4}{*}{Ultravox-v0.5-llama-3.1-8b} & \xmark & \xmark & 48.23${\pm2.09}$ & \textbf{\boldmath 52.33${\pm18.15}$} & 35.19${\pm0.69}$ & 27.67${\pm1.39}$ \\
 & \cmark & \xmark & 47.45${\pm0.52}$ & 21.54${\pm7.09}$ & 32.87${\pm0.49}$ & 22.08${\pm0.88}$ \\
 & \xmark & \cmark & 47.45${\pm0.98}$ & 28.34${\pm2.58}$ & 33.80${\pm0.45}$ & 25.81${\pm0.61}$ \\
 & \cmark & \cmark & 43.92${\pm1.05}$ & 38.82${\pm3.56}$ & 34.36${\pm1.56}$ & \textbf{\boldmath 31.84${\pm2.14}$} \\
\cmidrule(lr){1-7}
\multirow{4}{*}{SeaLLMs-Audio-7B} & \xmark & \xmark & 39.09${\pm7.76}$ & 25.79${\pm10.61}$ & 34.45${\pm1.03}$ & 25.12${\pm6.21}$ \\
 & \cmark & \xmark & 37.12${\pm2.66}$ & \underline{40.33${\pm15.03}$} & 34.59${\pm2.42}$ & 28.89${\pm2.01}$ \\
 & \xmark & \cmark & 48.37${\pm0.00}$ & 17.32${\pm2.40}$ & 33.39${\pm0.11}$ & 22.04${\pm0.63}$ \\
 & \cmark & \cmark & 43.79${\pm2.19}$ & 33.72${\pm5.18}$ & 34.30${\pm1.71}$ & 30.22${\pm3.28}$ \\
\cmidrule(lr){1-7}
\multirow{4}{*}{Qwen2-Audio-7B-Instruct} & \xmark & \xmark & 39.34${\pm3.64}$ & 23.37${\pm1.39}$ & 30.73${\pm0.55}$ & 25.28${\pm1.39}$ \\
 & \cmark & \xmark & 45.36${\pm0.67}$ & 24.26${\pm1.69}$ & 33.61${\pm1.00}$ & 26.17${\pm1.65}$ \\
 & \xmark & \cmark & 31.37${\pm2.15}$ & 20.29${\pm5.44}$ & 32.20${\pm0.62}$ & 19.86${\pm4.40}$ \\
 & \cmark & \cmark & 45.23${\pm3.95}$ & 38.50${\pm16.44}$ & 34.36${\pm1.95}$ & 27.85${\pm1.76}$ \\
\cmidrule(lr){1-7}
\multirow{4}{*}{MiniCPM-o.2.6} & \xmark & \xmark & 48.63${\pm0.67}$ & 32.63${\pm1.33}$ & 35.75${\pm1.58}$ & 27.68${\pm2.74}$ \\
 & \cmark & \xmark & \textbf{\boldmath 49.15${\pm0.49}$} & 39.69${\pm5.72}$ & 35.10${\pm0.69}$ & 25.76${\pm1.30}$ \\
 & \xmark & \cmark & \underline{48.76${\pm0.67}$} & 35.06${\pm7.73}$ & 34.76${\pm0.97}$ & 26.49${\pm2.33}$ \\
 & \cmark & \cmark & 48.63${\pm0.32}$ & 29.50${\pm16.38}$ & 33.73${\pm0.49}$ & 22.55${\pm1.01}$ \\
\cmidrule(lr){1-7}
\multirow{4}{*}{Phi-4-multimodal-instruct} & \xmark & \xmark & 45.88${\pm0.77}$ & 28.25${\pm1.06}$ & 37.11${\pm1.34}$ & \underline{31.03${\pm1.41}$} \\
 & \cmark & \xmark & 46.93${\pm0.96}$ & 28.88${\pm1.02}$ & 36.48${\pm1.64}$ & 29.83${\pm2.30}$ \\
 & \xmark & \cmark & 48.37${\pm0.00}$ & 32.78${\pm0.00}$ & 34.00${\pm0.22}$ & 23.85${\pm0.64}$ \\
 & \cmark & \cmark & 47.45${\pm1.06}$ & 29.18${\pm3.51}$ & 34.38${\pm0.93}$ & 26.07${\pm1.16}$ \\
\cmidrule(lr){1-7}
\multirow{4}{*}{Qwen2.5-Omni-3B} & \xmark & \xmark & 40.91${\pm7.12}$ & 26.15${\pm3.42}$ & 35.14${\pm1.91}$ & 28.07${\pm2.45}$ \\
 & \cmark & \xmark & 43.66${\pm2.63}$ & 25.94${\pm1.36}$ & 36.02${\pm1.13}$ & 29.58${\pm1.13}$ \\
 & \xmark & \cmark & 41.83${\pm8.13}$ & 28.01${\pm2.78}$ & 33.75${\pm1.89}$ & 26.64${\pm2.12}$ \\
 & \cmark & \cmark & 44.05${\pm0.89}$ & 25.66${\pm1.00}$ & 35.40${\pm1.59}$ & 29.03${\pm1.87}$ \\
\cmidrule(lr){1-7}
\multirow{4}{*}{Qwen2.5-Omni-7B} & \xmark & \xmark & 42.35${\pm4.56}$ & 26.68${\pm3.36}$ & 35.35${\pm0.87}$ & 28.52${\pm0.79}$ \\
 & \cmark & \xmark & 39.61${\pm2.60}$ & 25.57${\pm1.33}$ & \textbf{\boldmath 37.82${\pm1.92}$} & 29.93${\pm1.77}$ \\
 & \xmark & \cmark & 44.84${\pm1.41}$ & 26.59${\pm1.26}$ & 35.04${\pm0.56}$ & 28.70${\pm0.55}$ \\
 & \cmark & \cmark & 33.20${\pm2.28}$ & 31.94${\pm13.62}$ & \underline{37.82${\pm0.88}$} & 26.16${\pm1.90}$ \\
\bottomrule
\end{tabular}
\caption{Results of LLMs on the CIR-E dataset (\%).}
\label{tab:cir-e}
\end{table*}

\subsection{Case Study}
\begin{figure*}[h]
    \centering
    \includegraphics[width=0.95\linewidth]{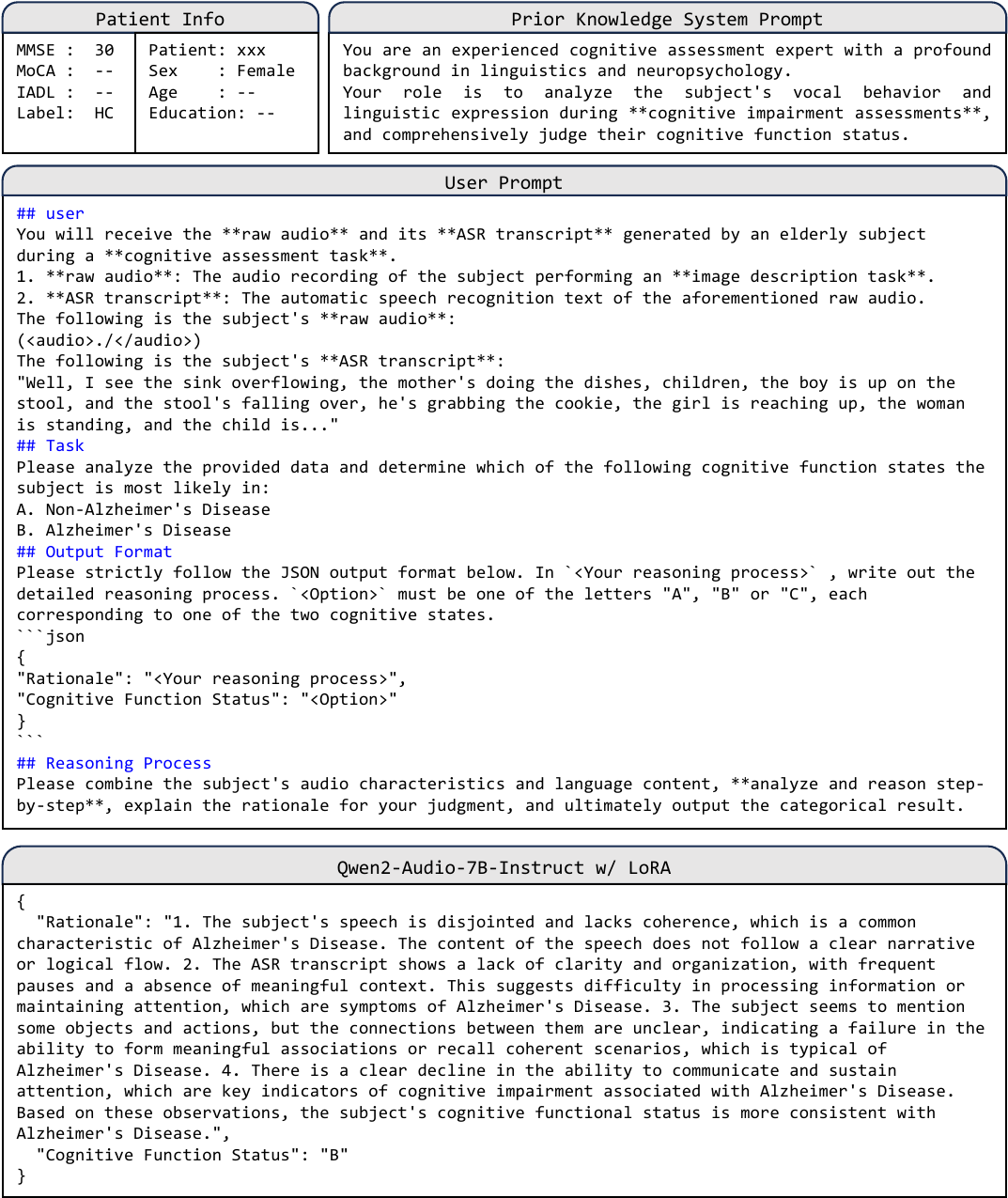}
    \caption{Illustrative example from Subject 1, belonging to the ADReSSo dataset.}
    \label{fig:case_1}
\end{figure*}

\begin{figure*}[h]
    \centering
    \includegraphics[width=0.95\linewidth]{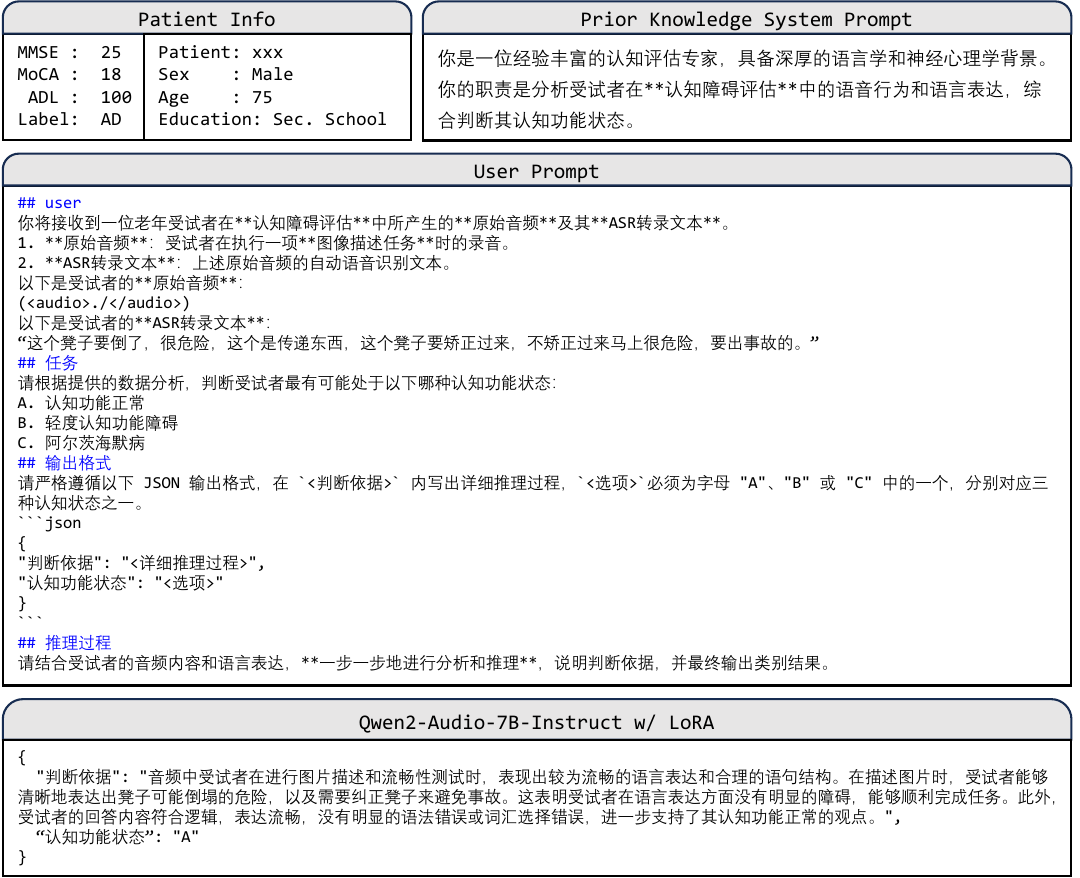}
    \caption{Illustrative example from Subject 2, belonging to the CIR-E dataset.}
    \label{fig:case_2}
\end{figure*}

The following sections present detailed case studies illustrating specific instances of model performance. Figures~\ref{fig:case_1} and~\ref{fig:case_2} showcase the particular situations of \textbf{Subject 1} and \textbf{Subject 2}, respectively, providing insights into the strengths and limitations of the LLM-based assessments.

\textbf{Subject 3:} This case involves a 66-year-old woman with a high school education. She scored $27/30$ on the MMSE and $26/30$ on the MoCA and was clinically judged cognitively normal. However, Qwen2-Audio repeatedly misdiagnosed her as AD or MCI.

\begin{figure*}[h]
    \centering
    \includegraphics[width=0.95\linewidth]{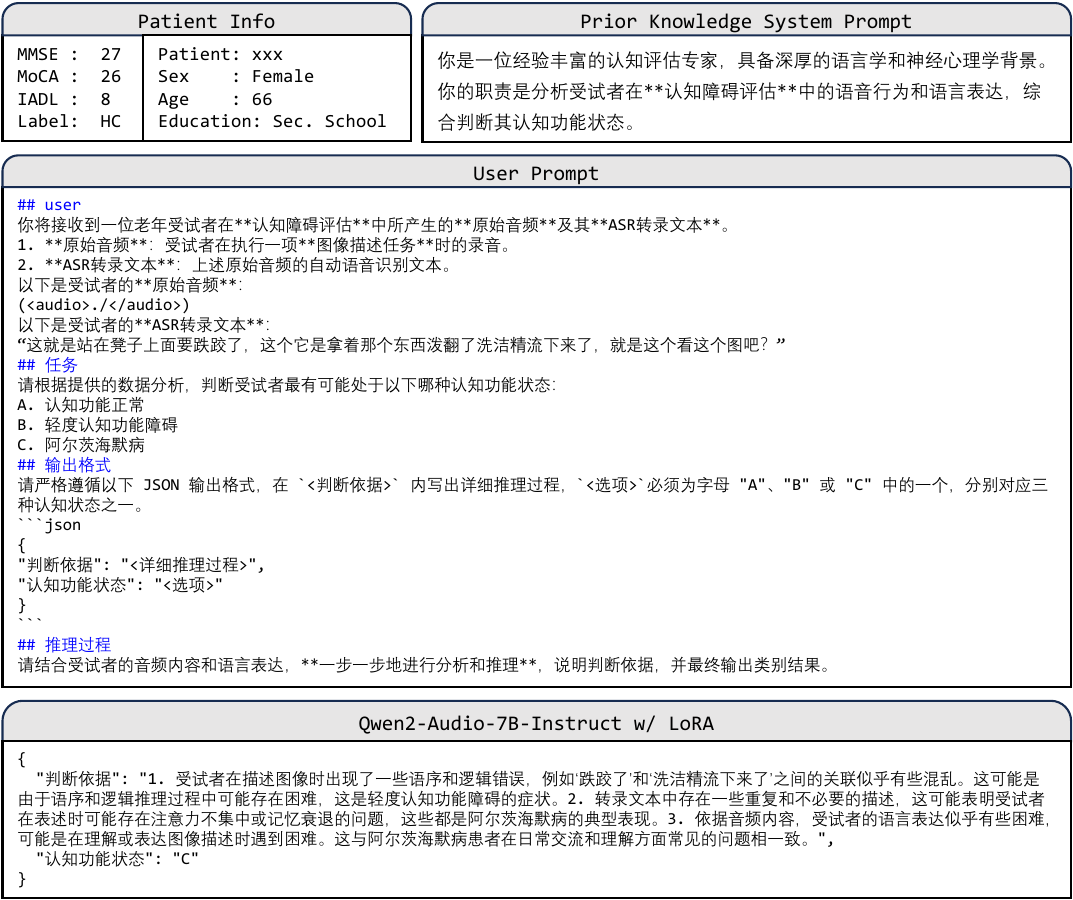}
    \caption{Illustrative example from Subject 3, belonging to the CIR-E dataset.}
    \label{fig:case_3}
\end{figure*}

The model’s CoT identified the subject’s “incoherent language and grammatical errors” as signs of AD, yet this interpretation contradicts her high scores on the MMSE and MoCA language subscales ($9/9$ and $5/6$). While her ASR transcript does contain some non-standard expressions (such as missing subjects or pronoun use), these are consistent with normal spoken language for someone of her age and educational background, and do not indicate disorganized thinking, as shown in Figure~\ref{fig:case_3}.
Additionally, the model interpreted her use of \textit{``that thing''} without a clear link to “detergent” as evidence of thought disorder. However, her slight loss of points on the MoCA naming subscale ($2/3$) more plausibly reflects mild, non-pathological word-finding difficulty, rather than a semantic breakdown due to AD. The model exaggerated this minor issue as indicative of widespread cognitive impairment.
Finally, the model attributed her closing confirmatory question (\textit{``This one, right? Should I look at this picture?''}) to memory decline. In reality, her scores on the MoCA delayed recall and MMSE recall subscales ($4/5$ and $2/3$) do not support significant memory impairment. Such questions are more likely to reflect task confirmation or a request for feedback.
In summary, without sufficient clinical and personal context, current LLMs risk over-interpreting normal variation as pathological features, acting as overly sensitive but insufficiently discerning “symptom detectors.”

\end{document}